\documentclass{article}

\usepackage{microtype}
\usepackage{graphicx}
\usepackage{subcaption}
\usepackage{booktabs}
\usepackage{hyperref}

\usepackage[acronym,nowarn]{glossaries}
\glsdisablehyper
\newacronym{llm}{LLM}{Large Language Model}
\newacronym{nlr}{NLR}{Natural Language Requirement}
\newacronym{sft}{SFT}{Supervised Fine-Tuning}

\usepackage[preprint]{icml2026}

\usepackage{amsmath}
\usepackage{amssymb}
\usepackage{mathtools}
\usepackage{amsthm}
\usepackage{listings}
\usepackage{tikz}
\usepackage{enumitem}
\usepackage{algorithm}
\usepackage{algorithmic}
\usepackage{xspace}
\usepackage{booktabs}
\usepackage{pifont}
\usepackage[table]{xcolor}
\usepackage{tcolorbox}
\tcbuselibrary{skins,listings,breakable,poster}
\usepackage{xparse}

\graphicspath{{figures/}}

\usepackage[capitalize,noabbrev]{cleveref}

\theoremstyle{plain}

\theoremstyle{definition}

\theoremstyle{remark}

\definecolor{keywordcolor}{rgb}{0.7, 0.1, 0.1}
\definecolor{commentcolor}{rgb}{0.4, 0.4, 0.4}
\definecolor{goodcolor}{RGB}{220,245,220}
\definecolor{badcolor}{RGB}{245,220,220}
\definecolor{goodcolor}{RGB}{200,235,200}

\lstdefinelanguage{lean}{
    morekeywords={def,theorem,lemma,example,structure,class,instance,where,if,then,else,match,with,let,in,have,show,by,sorry,Prop,Type,Nat,Int,List,Bool,true,false,reducible,simp},
    sensitive=true,
    morecomment=[l]{--},
    morecomment=[s]{/-}{-/},
    morestring=[b]",
    keywordstyle=\color{keywordcolor},
    commentstyle=\color{commentcolor},
}

\newcommand{\code}[1]{\lstinline|#1|}

\newif\ifcomments
\commentsfalse
\ifcomments
\newcommand{\todo}[1]{\textcolor{orange}{[TODO: #1]}}
\newcommand{\zhe}[1]{\textcolor{blue}{[zhe: #1]}}
\newcommand{\chloe}[1]{\textcolor{green}{[chloe: #1]}}
\newcommand{\soonho}[1]{\textcolor{teal}{[soonho: #1]}}
\newcommand{\aidan}[1]{\textcolor{yellow}{[aidan: #1]}}
\newcommand{\dawn}[1]{\textcolor{red}{[dawn: #1]}}
\newcommand{\ug}[1]{\textcolor{green}{[udaya: #1]}}
\else
\newcommand{\todo}[1]{}
\newcommand{\zhe}[1]{}
\newcommand{\chloe}[1]{}
\newcommand{\soonho}[1]{}
\newcommand{\dawn}[1]{}
\newcommand{\aidan}[1]{}
\newcommand{\ug}[1]{}
\fi

\newcommand{\framework}{\textsc{VeriSpecGen}\xspace}

\newcommand{\verina}{\textsc{Verina}\xspace}

\newcommand{\qwensmall}{Qwen3-4B-Instruct-2507\xspace}
\newcommand{\qwenmid}{Qwen3-Coder-30B-A3B\xspace}
\newcommand{\qwenlarge}{Qwen3-Coder-480B-A35B\xspace}
\definecolor{darkgreen}{rgb}{0,0.5,0}
\newcommand{\smallsec}[1]{\textbf{#1}}

\icmltitlerunning{Intent-aligned Formal Specification Synthesis via Traceable Refinement}

\begin{document}

\twocolumn[
  \icmltitle{Intent-aligned Formal Specification Synthesis via Traceable Refinement}

  \icmlsetsymbol{equal}{*}

  \begin{icmlauthorlist}
    \icmlauthor{Zhe Ye$^\dagger$}{ber}
    \icmlauthor{Aidan Z.H. Yang}{aws}
    \icmlauthor{Huangyuan Su$^\dagger$}{har}
    \icmlauthor{Zhenyu Liao}{aws}
    \icmlauthor{Samuel Tenka}{aws}
    \icmlauthor{Zhizhen Qin}{aws}
    \icmlauthor{Udaya Ghai}{aws}
    \icmlauthor{Dawn Song}{ber}
    \icmlauthor{Soonho Kong}{aws}
  \end{icmlauthorlist}

  \icmlaffiliation{ber}{UC Berkeley}
  \icmlaffiliation{aws}{Amazon Web Services}
  \icmlaffiliation{har}{Harvard University}

  \icmlcorrespondingauthor{Soonho Kong}{soonho@amazon.com}

  \icmlkeywords{Formal Verification, Specification Synthesis, Large Language Models, Lean 4}

  \vskip 0.3in
]

\printAffiliationsAndNotice{$^\dagger$Work done during internship at Amazon Web Services.}


\begin{abstract}
Large language models are increasingly used to generate code from natural language, but ensuring correctness remains challenging.
Formal verification offers a principled way to obtain such guarantees by proving that a program satisfies a formal specification.
However, specifications are frequently missing in real-world codebases, and writing high-quality specifications remains expensive and expertise-intensive.
We present \framework{}, a traceable refinement framework that synthesizes intent-aligned specifications in Lean through requirement-level attribution and localized repair.
\framework{} decomposes natural language into atomic requirements and generates requirement-targeted tests with explicit traceability maps to validate generated specifications.
When validation fails, traceability maps attribute failures to specific requirements, enabling targeted clause-level repairs.
\framework{} achieve 86.6\% on \verina{} SpecGen task using Claude Opus 4.5, improving over baselines by up to 31.8 points across different model families and scales.
Beyond inference-time gains, we generate 343K training examples from \framework{} refinement trajectories and demonstrate that training on these trajectories substantially improves specification synthesis by 62--106\% relative and transfers gains to general reasoning abilities.
\end{abstract}

\section{Introduction}
\label{sec:intro}

Large language models (LLMs) are increasingly used to generate code from natural language instructions~\cite{chen2021evaluating,roziere2023code}, enabling rapid implementation while introducing new challenges for correctness assurance.
Recent studies have shown that LLM-generated code frequently contains functional errors~\cite{wang2025towards} and security vulnerabilities~\cite{guo2024redcode,yang2024seccodeplt}\dawn{you can add RedCode and SecCodePLT etc. for citations as well}\zhe{added}, and these issues typically require costly human review to resolve.
Formal verification offers a principled approach to automated correctness assurance by proving that a program satisfies a \emph{formal specification}---a machine-checkable behavioral contract usually expressed as preconditions and postconditions~\cite{hoare1969axiomatic}. 
Recent work demonstrates that LLMs can assist verification workflows by generating specifications or proofs~\cite{sun2024clover,misu2024towards,aggarwal2024alphaverus,ye2025verina}, suggesting the potential for end-to-end verifiable code generation.

However, the strength of verification guarantees is fundamentally bounded by specification quality, yet synthesizing correct specifications remains challenging even for human experts~\cite{woodcock2009formal}.
First, specifications require precise intent alignment between informal descriptions and formal contracts.
A specification that is too weak permits buggy implementations to pass verification, while an incorrect specification rejects correct implementations, undermining the value of verification in either case.
Achieving this alignment demands translating potentially ambiguous natural language into unambiguous preconditions and postconditions that capture exactly the intended behavior.
Second, training data for specification synthesis is scarce because formal specifications are rarely present in real-world codebases and remain expensive to write.
Recent benchmarks show that even state-of-the-art models struggle to reliably synthesize specifications~\cite{sun2024clover, ye2025verina,thakur2025clever}, demonstrating that specification generation remains a substantial bottleneck for verifiable code generation.
Existing approaches to specification synthesis, as discussed in~\Cref{sec:background}, either anchor specification correctness on program artifacts~\cite{ernst2007daikon,ma2024specgen,jin2024autospec,sun2025classinvgen} or fail to effectively repair misaligned specifications~\cite{ghosh2016arsenal,cosler2023nl2spec,endres2024can,sun2024clover,cao2025informal}.

\begin{figure*}[t]
    \centering
    \includegraphics[width=0.85\linewidth]{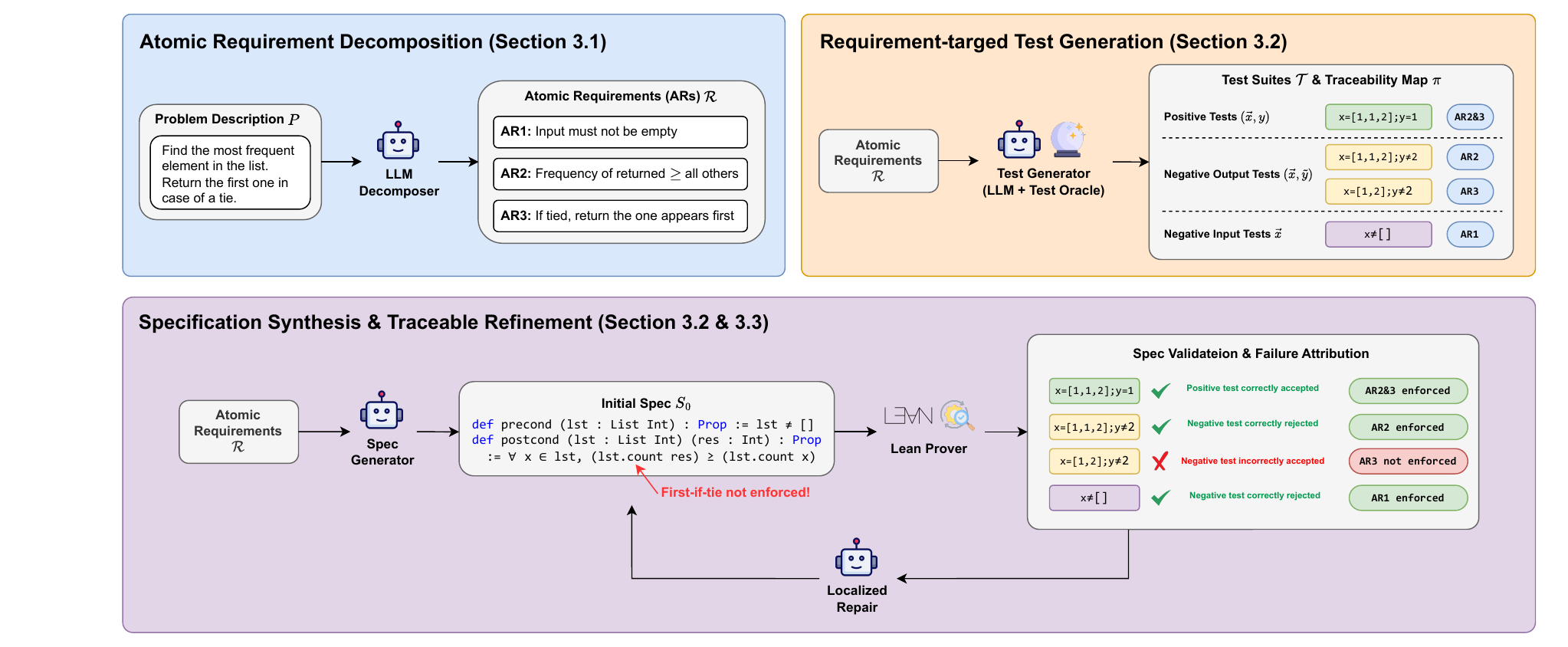}
    \caption{
    \textbf{\framework{} traceable refinement workflow.} Given a natural language problem description (e.g., ``Find the most frequent element...''), \framework{} synthesizes formal specifications through three stages: (1) \textbf{Atomic Requirement Decomposition}: An LLM decomposes the description into testable atomic requirements (AR1: non-empty input, AR2: frequency constraint, AR3: tie-breaking rule). (2) \textbf{Requirement-targeted Test Generation}: For each requirement, the system generates positive tests (valid input-output pairs), negative-output tests (invalid outputs for valid inputs), and negative-input tests (invalid inputs), creating an explicit traceability map $\pi$ linking each test to its validating requirements. (3) \textbf{Specification Synthesis \& Traceable Refinement}: An LLM generates an initial Lean specification from the atomic requirements. A Lean prover validates the specification against all tests. When validation fails (e.g., test $\langle[1,2], y{\neq}2\rangle$ incorrectly accepted), the traceability map attributes the failure to specific requirements (AR3 via $\pi$), enabling \textbf{localized repair} that modifies only the affected postcondition clause rather than rewriting the entire specification. This process iterates until all tests pass, producing intent-aligned specifications.
    }
    \label{fig:workflow}
    \vspace{-5mm}
\end{figure*}

We address this challenge with \framework{}, a traceable refinement framework that synthesizes intent-aligned formal specifications in Lean~\cite{de2015lean} through requirement-level attribution and localized repair.
Given a natural language problem description, \framework{} decomposes it into informal atomic requirements and generates requirement-targeted test cases for each requirement, including positive examples that should satisfy the specification and negative examples that should be rejected.
Test outputs are labeled by executing a reference implementation as a black-box oracle, without inspecting code structure or deriving specification content from the implementation.
Each test is paired with an explicit traceability map that links back to the requirements it validates.
\framework{} generates an initial specification from the atomic requirements and validates it by instantiating each test into a Lean proposition and attempting to prove or disprove it using proof automation.
When validation fails, the traceability map attributes failures to specific requirements, enabling localized clause-level repairs that modify only the affected contract components.
This process iterates until all tests pass and adversarial testing does not reveal missing constraints, yielding specifications that align with natural language intent through targeted, requirement-grounded refinement.

To address the scarcity of training data for specification synthesis, we generate a large-scale dataset from \framework{}'s traceable refinement process.
We execute \framework{} with Claude Sonnet 4.5~\cite{anthropic2025sonnet45} on TACO-verified~\cite{likaixin2024taco-verified}, producing structured trajectories containing requirement decompositions, traceability maps, intermediate specifications, and localized repairs.
We distill these trajectories into supervised fine-tuning examples, yielding 343,827 high-quality instruction-response pairs.

We evaluate \framework{} on the SpecGen task of the \verina{} benchmark.
Across multiple model scales, traceable refinement improves \verina{} SpecGen scores by up to 31.8 points over the benchmark baselines, reaching 86.6 points with Claude Opus 4.5~\cite{anthropic2025opus45} and achieving state-of-the-art performance.
Ablation studies show that all framework components are necessary for effective refinement, with requirement decomposition as the most critical component.
Beyond inference-time improvements, we demonstrate that training on trajectory-distilled data substantially improves base model capabilities.
Fine-tuning \qwensmall and \qwenmid~\cite{yang2025qwen3} on our dataset improves both \verina{} SpecGen scores by 15.6--18.2 points (62--106\% relative) and \verina{} CodeGen scores by 18.2--21.5 points (54--72\% relative), and transfers improvements to out-of-domain math reasoning and general coding benchmarks.

In summary, our contributions are:
\begin{itemize}[leftmargin=1.2em, itemsep=0pt]
\item \textbf{Traceable refinement for specification synthesis.}
We introduce traceable refinement that enables targeted, localized specification repairs.
By explicitly mapping validation failures to specific natural language requirements, \framework{} achieves state-of-the-art specification synthesis, improving over direct generation baselines up to 31.8 percentage points across model scales.
\item \textbf{Scalable dataset generation via trajectory distillation.}
We generate the first large-scale and high-quality dataset of traceable refinement trajectories with 343K examples.
Training on these trajectories improves specification synthesis by up to 106\% relatively while transferring to out-of-domain reasoning tasks, providing a scalable path to improve specification synthesis models.
\end{itemize}


\section{Background and Related Works}
\label{sec:background}

We discuss works related to \framework{} in detail below.

\smallsec{Specification synthesis paradigms.}
Specification synthesis approaches fall into two paradigms.
Program-behavior-based methods infer specifications from code via dynamic analysis~\cite{ernst2007daikon} or LLMs~\cite{ma2024specgen,wen2024enchanting,sun2025classinvgen}, excelling when implementations are trusted ground truth.
Intent-anchored methods synthesize specifications from natural language~\cite{ghosh2016arsenal,cosler2023nl2spec,endres2024can,sun2024clover,mugnier2025laurel,mukherjee2024towards,cao2025informal}, matching workflows where LLM-generated code may be buggy.
However, prior intent-anchored systems cannot determine whether specifications capture all intended requirements and rely on coarse refinement signals (type errors, proof failures) that cannot localize failures to specific requirements.
\framework{} follows the intent-anchored paradigm, deriving specification content from natural language while using reference implementations solely as black-box oracles for test labeling rather than extracting specification logic.
\framework{} addresses the refinement gap through traceable refinement: decomposing natural language into atomic requirements, generating requirement-targeted tests, and using traceability maps to enable requirement-level attribution and localized repairs when validation fails.


\smallsec{Refinement mechanisms and trajectory-based learning.}
Most prior work employs coarse refinement signals such as verifier errors or global test outcomes~\cite{madaan2023self,chen2023teaching} that indicate failure without localizing the source of misalignment.
Fine-grained signals like mutation testing~\cite{sun2025classinvgen} provide counterexamples but do not map directly to requirements in the natural-language description.
Recent work demonstrates the effectiveness of learning from refinement trajectories for general reasoning tasks~\cite{zelikman2022star,chen2023fireact,shinn2023reflexion} and formal artifact generation~\cite{lin2024lean}.
To the best of our knowledge, \framework{} is the first to provide requirement-level attribution through requirement decomposition, enabling targeted clause-level repairs.
\framework{} further generates a large-scale trajectory dataset for training purpose that provides process-level supervision for achieving intent alignment.


\section{Traceable Refinement for Specification Synthesis}
\label{sec:method}

We propose \framework{}, an agentic framework for synthesizing intent-aligned formal specifications through traceable refinement.
The workflow is illustrated in~\Cref{fig:workflow}.

\smallsec{Task definition.}
Given a natural-language problem description $P$ and a function signature $\sigma$ that defines the input/output types, our goal is to synthesize an intent-aligned Lean~4 specification $S=\langle\mathsf{pre}(\vec{x}),\mathsf{post}(\vec{x}, y)\rangle$.
The synthesized precondition $\mathsf{pre}(\vec{x})$ constrains the set of valid inputs $\vec{x}$ and should properly reject invalid inputs according to $P$.
The postcondition $\mathsf{post}(\vec{x}, y)$ characterizes the set of correct outputs $y$ for each valid input and should properly reject undesired outputs according to $P$.

\smallsec{Framework overview.}
Specification synthesis requires high-quality refinement signals to guide iterative improvement, but such signals are difficult to obtain.
Type checking and proof failures provide only coarse feedback indicating global inconsistency, while direct LLM-based judging on specification is unreliable and may hallucinate alignment.

\framework{} addresses this through \emph{traceable refinement}.
We decompose the natural-language description into \emph{atomic requirements}, which are informal statements each expressing a single testable behavioral property (\Cref{sec:decomp}).
For each requirement, we generate targeted test cases that establish an explicit \emph{traceability map} linking tests to requirements (\Cref{sec:testgen}).
When tests fail, we trace failures back to implicated requirements and perform localized repairs on the specification (\Cref{sec:repair}).
This process continues until all tests pass, followed by adversarial testing to expose missing constraints.
During refinements, we collect the trajectories for fine-tuning dataset construction (\Cref{sec:trace}).

\dawn{in general in this section, it's helpful to give some concrete examples, and refer to Figure 1 often, so it's easier for readers to understand.}

\subsection{Decomposition into Atomic Requirements}
\label{sec:decomp}

To enable fine-grained refinement feedback, \framework{} decomposes the natural-language description $P$ into a set of atomic requirements (ARs) $\mathcal{R}=\{r_i\}_{i=1}^m$ using an LLM.
An atomic requirement $r_i$ is an informal natural-language statement expressing a \emph{single} behavioral property that can be validated on concrete input-output examples, rather than bundling multiple properties in one requirement.
We prompt the model to extract such requirements from $P$, explicitly requesting both functional requirements and edge-case requirements including boundary conditions and tie-breaking rules.
\framework{} further leverages an LLM-based judge to review and improve $\mathcal{R}$. The resulting requirement set $\mathcal{R}$ serves as the unit of traceability for test generation (\Cref{sec:testgen}) and failure attribution (\Cref{sec:repair}).
For example in~\Cref{fig:workflow}, the example description decomposes into three atomic requirements.

\subsection{Requirement-Targeted Tests Generation and Specification Validation}
\label{sec:testgen}

\framework{} uses requirement-targeted tests as \emph{grounded} validation signals for a candidate specification $S$, instead of purely subjective LLM judgment directly.
For each requirement $r\in\mathcal{R}$, we construct a set of labeled test cases intended to validate the behavioral property $r$.
Concretely, we generate three types of tests: (i) \emph{positive tests} $(\vec{x}, y)$ that should be accepted by the candidate specification, (ii) \emph{negative-input tests} $\vec{x}$ that should be rejected by the candidate precondition, and (iii) \emph{negative-output tests} $(\vec{x}, \tilde{y})$ that should be rejected by the candidate postcondition for valid inputs.

\smallsec{Test oracles for output labeling.}
To label test outputs, we require a test oracle $\mathcal{O}$ that provides ground-truth outputs for given inputs.
We instantiate $\mathcal{O}$ as a black-box execution oracle using a reference implementation, either provided with $P$ or generated by an LLM.
$\mathcal{O}$ is agnostic to individual atomic requirements.
Mining specifications directly from implementations (e.g., via symbolic execution or program analysis) would anchor validation to potentially buggy code rather than the user's intent.
Instead, we use the oracle only for labeling concrete test outputs, not for extracting specifications, and therefore maintain intent-anchored validation where specification correctness is judged against natural-language requirements.

\smallsec{Test generation.}
The inputs $\vec{x}$ for positive and negative-input tests are generated by prompting an LLM with the specific requirement $r$ to match its intent, emphasizing boundary values and adversarial cases.
To label outputs for positive and negative-output tests, we query the oracle $y:=\mathcal{O}(\vec{x})$ and perturb $y$ into different $\tilde{y}$s using an LLM to violate requirement-relevant constraints.
We collect all tests into a test suite $\mathcal{T}$ and annotate each test $t\in\mathcal{T}$ with its target requirement(s) via a traceability map $\pi:\mathcal{T}\rightarrow 2^{\mathcal{R}}$, enabling requirement-level attribution of failing validations during refinement.
For instance, in~\Cref{fig:workflow}, test$\langle[1,2],2\rangle$ maps to AR3, enabling failure attribution.

\smallsec{Validating specifications using tests.}
Given a candidate specification $S$ and a test suite $\mathcal{T}$, we must determine whether $S$ correctly enforces each requirement.
Following \verina{}~\cite{ye2025verina}, we construct a Lean proposition for each test $t\in\mathcal{T}$ by instantiating $S$ with the concrete values from $t$, and attempt to prove or disprove it using proof automation such as the \texttt{grind} tactic~\cite{lean4grind}.
We use proof automation rather than execution because specifications are logical predicates, not executable functions, and the prover can check whether the proposition is logically entailed by the specification.

For a candidate specification $S$, we define three check propositions $\phi^{+}$, $\phi^{\text{in}-}$, and $\phi^{\text{out}-} $ based on test type:
\begin{itemize}[leftmargin=1em, itemsep=0pt]
\item \textbf{Positive test} $\phi^{+}(\vec{x}, y)$: $\phi^{+}(S, t):= \mathsf{pre}(\vec{x}) \wedge \mathsf{post}(\vec{x}, y)$, which checks that the specification accepts an intended input-output pair.
\item \textbf{Negative-input test} $\vec{x}$: $\phi^{\text{in}-}(S, t) := \neg \mathsf{pre}(\vec{x})$, which checks that the precondition rejects an invalid input.
\item \textbf{Negative-output test} $(\vec{x}, \tilde{y})$: $\phi^{\text{out}-}(S, t) := \mathsf{pre}(\vec{x}) \wedge \neg \mathsf{post}(\vec{x}, \tilde{y})$, which checks that the postcondition rejects an incorrect output for a valid input.
\end{itemize}

We define $t$ as a \texttt{pass} on $S$ if Lean automation can prove the instantiated proposition $\phi(S, t)$, and a \texttt{fail} if Lean proof automation proves its negation.
If automation is inconclusive, we invoke an LLM judge \emph{only for that test} to label the concrete test as \texttt{pass} or \texttt{fail}.
The judge is used purely to guide refinement in case proof automation fails; after each update we re-run Lean on the full suite.

We summarize the validation results using an evaluation function $\textsc{eval}(S,\mathcal{T})$ that partitions tests by their outcomes:
\begin{align}
\label{eq:eval}
\begin{split}
&\textsc{eval}(S,\mathcal{T})
:= \\
&\big\langle
\mathcal{T}^{\mathrm{Lean}}_{\textsc{pass}}(S),\;
\mathcal{T}^{\mathrm{Lean}}_{\textsc{fail}}(S),\;
\mathcal{T}^{\mathrm{Judge}}_{\textsc{pass}}(S),\;
\mathcal{T}^{\mathrm{Judge}}_{\textsc{fail}}(S)
\big\rangle
\end{split}
\end{align}
where $\mathcal{T}^{\mathrm{Lean}}_{\textsc{pass}}(S)$ and $\mathcal{T}^{\mathrm{Lean}}_{\textsc{fail}}(S)$ are tests whose corresponding propositions are proved true or false by Lean.
$\mathcal{T}^{\mathrm{Judge}}_{\textsc{pass}}(S)$ and $\mathcal{T}^{\mathrm{Judge}}_{\textsc{fail}}(S)$ are the remaining tests labeled by the judge due to proof automation being inconclusive.

For refinement, we use the failing tests
\begin{equation}
\label{eq:failing-tests}
\mathcal{F}(S)
:= \mathcal{T}^{\mathrm{Lean}}_{\textsc{fail}}(S)\ \cup\ \mathcal{T}^{\mathrm{Judge}}_{\textsc{fail}}(S)
\end{equation}
and lift them to linked requirements via traceability map $\pi$:
\begin{equation}
\label{eq:failing-reqs}
\mathcal{R}_{\textsc{fail}}(S)
:= \bigcup\nolimits_{t\in\mathcal{F}(S)} \pi(t)
\end{equation}
$\mathcal{R}_{\textsc{fail}}(S)$ identifies which natural-language requirements are implicated by validation failures, enabling requirement-localized clause repair in \Cref{sec:repair}.

\subsection{Specification Synthesis and Requirement-Localized Repair}
\label{sec:repair}

\smallsec{Iterative refinement via requirement-localized feedback.}
We first generate an initial specification $S_0$ from the atomic requirement set $\mathcal{R}$ by prompting an LLM to translate $\mathcal{R}$ into Lean~4 preconditions and postconditions.
The LLM is prompted to structure the specification such that individual requirements correspond to separate clauses.

Given candidate specification $S_k$ at iteration $k$, we evaluate it on test suite $\mathcal{T}$ (\Cref{eq:eval}) to obtain failing tests $\mathcal{F}(S_k)$ (\Cref{eq:failing-tests}) and implicated requirements $\mathcal{R}_{\textsc{fail}}(S_k)$ (\Cref{eq:failing-reqs}).
The traceability map $\pi$ attributes each test failure to specific requirements, enabling targeted feedback.
\dawn{note that the tracability map is not entirely accurate; may need to make this clear.}\zhe{clarified in 3.2 test generation}

We construct a repair prompt containing the current specification $S_k$, the implicated failing requirements, and their corresponding representative failing tests.
The LLM generates an updated specification $S_{k+1}$ that addresses the implicated requirements while preserving correct aspects of $S_k$.
Iteration continues until all tests pass or a maximum iteration budget is exhausted.
As an example, in~\Cref{fig:workflow}, the initial specification fails negative output test $\langle[1,2],2\rangle{}$, which maps to AR3 via $\pi$, triggering localized repair of the postcondition to enforce the tie-breaking constraint.

\smallsec{Adversarial testing.}
When a candidate specification $S_k$ passes all tests in $\mathcal{T}$ (i.e., $\mathcal{F}(S_k)=\emptyset$), we further apply adversarial testing to detect incorrect or under-constrained specifications.
We prompt an LLM to generate adversarial test cases in all three test types based on the atomic requirements $\mathcal{R}$ and are designed to break the current formal specification $S_k$.
These adversarial tests aim to expose edge cases, missing constraints, or overly permissive specifications that allow unintended behavior.
For each generated adversarial test $t_{\text{adv}}$, we validate $S_k$ on it using the same validation procedure as in \Cref{sec:testgen}.
If the adversarial test successfully breaks the specification (i.e., $S_k$ incorrectly accepts or rejects the test), we add $t_{\text{adv}}$ to the test suite $\mathcal{T}$ and use an LLM to map it to its target requirements via the traceability map $\pi$.
The newly discovered failures trigger additional refinement iterations, returning to the requirement-localized repair process with the expanded test suite.

\subsection{Trajectory Recording}
\label{sec:trace}
We record the complete synthesis trajectory:
\begin{equation}
\tau = (P,\sigma,\mathcal{R},\mathcal{T},\pi,S_0,\mathcal{F}_0,\Delta_0,S_1,\mathcal{F}_1,\Delta_1,\dots,S_K)
\end{equation}
where $\Delta_k$ denotes the localized feedback provided at iteration $k$ containing the implicated requirements and their associated failing tests.
These trajectories provide process-level supervision that bridges informal intent and formal contracts through intermediate reasoning steps.
Unlike end-to-end specification examples, trajectories expose how to achieve intent alignment: the decomposition from $P$ to $\mathcal{R}$ demonstrates requirement extraction, the traceability map $\pi$ provides requirement-to-test alignment, and the repair sequence $\langle S_k,\mathcal{F}(S_k),\Delta_k \rangle$ shows how to perform localized repair.
We leverage these trajectories for fine-tuning data construction detailed in \Cref{sec:dataset}.


\section{Training Data Construction via Trajectory Distillation}
\label{sec:dataset}

\begin{table}[t]
\small
\centering
\caption{
\textbf{Dataset statistics from trajectory distillation.}
Starting from 12,901 TACO-verified problems (decontaminated with zero 10-gram overlap with \verina{}), we successfully translate 11,981 to Lean and synthesize 6,842 specifications, from which we distill three SFT dataset variants.
}
\label{tab:dataset_stats}
\vspace{-2mm}
\begin{tabular}{lr}
\toprule
\textbf{Dataset} & \textbf{\#Instances} \\
\midrule
\multicolumn{2}{l}{\textit{Source and Processing}} \\
TACO-verified (decontaminated) & 12{,}901 \\
Translated to Lean with test suites & 11{,}981 \\
Successfully synthesized specifications & 6{,}842 \\
\midrule
\multicolumn{2}{l}{\textit{Distilled Training Examples}} \\
\textit{SFT Full} & 343{,}827 \\
\textit{SFT No-Test}  & 121{,}531 \\
\textit{SFT Spec-Only}  & 6{,}842 \\
\bottomrule
\end{tabular}
\vspace{-5mm}
\end{table}

High-quality training data for specification synthesis is scarce.
Formal specifications rarely exist in real-world codebases, and writing intent-aligned contracts remains expensive even for experts.
To improve model performance beyond inference-time techniques, we need scalable methods to synthetically generate specification synthesis data.

Prior work treats specification generation as direct natural-language-to-specification translation~\cite{endres2024can, cao2025informal}, providing problem descriptions paired with ground-truth contract labels.
While this establishes correctness targets, it offers no supervision for \emph{how} to achieve intent alignment.
These datasets lack guidance on which requirements to extract, how to validate alignment, or how to repair mis-specifications when validation fails.

\framework{}'s agentic workflow naturally produces this missing supervision.
Each synthesis trajectory contains intermediate reasoning steps such as requirement decompositions, traceability mappings, and failure attributions that bridge informal intent and formal contracts.
We leverage this structure through \emph{agentic trajectory distillation}~\cite{zelikman2022star,chen2023fireact,shinn2023reflexion}, executing \framework{} on Python problems from TACO-Verified to construct supervised training data.
Each trajectory is distilled into task-specific training examples for decomposition, validation, attribution, and targeted repair.
This provides process-level supervision beyond end-to-end examples.
We use Claude Sonnet 4.5 as the teacher model to ensure dataset quality, given its adequate performance on \verina{} SpecGen task.
Table~\ref{tab:dataset_stats} summarizes the statistics of dataset construction pipeline.

\smallsec{Source data and Lean translation.}
We source problems from TACO-Verified~\cite{likaixin2024taco-verified}, an MIT-licensed collection of Python competitive programming problems with reference solutions and tests.
We verify \textit{zero} 10-gram overlap with \verina{} to prevent evaluation leakage.
For each problem, we use the teacher model to generate the Lean signature from its Python reference solution and construct a literal translator to convert Python test values to type-correct Lean literals.
We successfully process 11,981 of 12,901 problems.

\smallsec{Trajectory collection.}
We execute \framework{} on each translated problem, using the problem description and Lean signature as inputs.
After decomposing requirements, we reuse the translated Python test suites as positive tests and map them to atomic requirements using the teacher model.
We prune tests based on requirement coverage to control test suite size.
Following \framework{}'s workflow, we then construct negative tests and execute specification refinement for up to 10 iterations per problem.
Each execution produces a comprehensive test suite including adversarial tests, a complete refinement trace, and a final specification when synthesis succeeds.
Out of 11,981 problems, we successfully synthesize 6,842 specifications.

\smallsec{SFT dataset distillation.}
We distill each trajectory checkpoint from the 6,842 successful synthesis runs into instruction-response pairs aligned with \framework{}'s modular components.
We define 11 tasks spanning the full synthesis pipeline, including requirement decomposition, test generation and mapping, specification generation, validation, failure attribution, and localized repair.
Each training example contains instructions based on its task with the teacher model's output as the target.
For instance, repair examples include failing requirements, failing tests, and the current specification.
We further construct three SFT dataset variants.
\textit{SFT Full} contains 343,827 examples spanning all tasks.
\textit{SFT No-Test} contains 121,531 examples, excluding test construction and mapping tasks.
\textit{SFT Spec-Only} contains 6,842 end-to-end NL-to-specification pairs.


\section{Evaluation}
\label{sec:eval}

To validate our approach, we measure both inference-time gains from traceable refinement and whether the refinement process itself can be distilled into effective training supervision by investigating three research questions:
\begin{description}[leftmargin=0em, itemsep=2pt, parsep=0pt]
\item[RQ1] Does traceable refinement improve specification synthesis compared to direct generation? We compare \framework{} against baselines across multiple model scales. (\Cref{sec:rq1})
    
\item[RQ2] Which components of traceable refinement are necessary for effective refinement? We isolate the contribution of requirement decomposition, traceability mapping, and adversarial testing through systematic ablations. (\Cref{sec:rq2})
    
\item[RQ3] Can refinement trajectories be distilled into training data that improves base model capabilities? We investigate whether structured trajectories provide richer supervision than end-to-end specification examples, enabling smaller models to internalize refinement strategies and generalize to out-of-domain tasks. (\Cref{sec:rq3})
\end{description}

\smallsec{General setup.}
For all experiments, we use Lean v4.24.0 with the \texttt{grind} tactic for proof automation (120s timeout).
For evaluation on \verina{}~\cite{ye2025verina}, we use the official harness and report both CodeGen and SpecGen pass@1 scores.
Unless otherwise specified, we use temperature $T=0.3$ and a maximum of 10,000 tokens for all LLM generation.
All inference and training are performed on AWS p5en.24xlarge instances with 8$\times$H200 GPUs, 192 CPU cores, and 2TB memory.

\subsection{RQ1: Does Traceable Refinement Improve Specification Synthesis?}
\label{sec:rq1}

\begin{table}[t]
\small
\centering
\caption{
\textbf{Specification synthesis results on \verina{}.}
\framework{} consistently outperforms \verina{} baseline across all models, achieving 42--87\% relative improvement.
}
\label{tab:verina_main}
\vspace{-2mm}
\resizebox{\linewidth}{!}{
\begin{tabular}{lccc}
\toprule
\textbf{Model} & \textbf{Baseline} & \textbf{\framework{}} & \textbf{$\Delta$ Abs. (Rel.)} \\
\midrule
\multicolumn{4}{l}{\small\textsc{Closed-Source Models}} \\
\midrule
Claude Opus 4.5   & 59.0 & \textbf{86.6} & \textcolor{darkgreen}{$\uparrow$27.6 (+46.8\%)} \\
Claude Sonnet 4.5 & 50.7 & 72.0 & \textcolor{darkgreen}{$\uparrow$21.3 (+42.0\%)} \\
\midrule
\multicolumn{4}{l}{\small\textsc{Open-Weight Models}} \\
\midrule
\qwenlarge   & 36.7 & 68.5 & \textcolor{darkgreen}{$\uparrow$31.8 (+86.7\%)} \\
\qwenmid    & 29.3 & 50.3 & \textcolor{darkgreen}{$\uparrow$21.0 (+71.7\%)} \\
\qwensmall   & 15.8 & 23.0 & \textcolor{darkgreen}{$\uparrow$7.2 (+45.6\%)} \\
\bottomrule
\end{tabular}
}
\vspace{-2mm}
\end{table}

\begin{table}[t]
\small
\centering
\caption{
\textbf{Ablation study on \verina{}.}
All variants improve over direct generation on Opus 4.5, and removing the decomposer causes the largest performance drop, showing that decomposition is critical for effective specification synthesis.
}
\label{tab:verina_ablation}
\vspace{-2mm}
\resizebox{\linewidth}{!}{
\begin{tabular}{lcccc}
\toprule
& \multicolumn{2}{c}{\textbf{Claude Opus 4.5}} & \multicolumn{2}{c}{\textbf{\qwenmid}} \\
\cmidrule(lr){2-3}\cmidrule(lr){4-5}
\textbf{Variant} & \textbf{Score} & \textbf{$\Delta$ Abs. (Rel.)} & \textbf{Score} & \textbf{$\Delta$ Abs. (Rel.)} \\
\midrule
Direct generation & 59.0 & -- & 29.3 & -- \\
\midrule
Direct refinement & 72.2 & \textcolor{darkgreen}{$\uparrow$13.2 (+22.4\%)} & 27.0 & \textcolor{red}{$\downarrow$2.3 (-7.8\%)} \\
Test-driven refinement & 73.8 & \textcolor{darkgreen}{$\uparrow$14.8 (+25.1\%)} & 20.9 & \textcolor{red}{$\downarrow$8.4 (-28.7\%)} \\
w/o adversarial testing & 81.0 & \textcolor{darkgreen}{$\uparrow$22.0 (+37.3\%)} & 48.9 & \textcolor{darkgreen}{$\uparrow$19.6 (+66.9\%)} \\
\midrule
\textbf{\framework{}} & \textbf{86.6} & \textcolor{darkgreen}{$\uparrow$27.6 (+46.8\%)} & \textbf{50.3} & \textcolor{darkgreen}{$\uparrow$21.0 (+71.7\%)} \\
\bottomrule
\end{tabular}
}
\vspace{-6mm}
\end{table}

\begin{table*}[h]
\small
\centering
\caption{
\textbf{Evaluation of trajectory-distilled training variants on \verina{}}.
Training on specification synthesis trajectories improves both CodeGen and SpecGen tasks.
\textit{SFT No-Test} achieves the best specification performance (+15.6--18.2 SpecGen@1), while \textit{SFT Full} yields the strongest code generation transfer (+18.2--21.5 CodeGen@1).
$\Delta$ shows both absolute and relative improvements over the base model.
}
\label{tab:rq3_training}
\vspace{-2mm}
\resizebox{\textwidth}{!}{
\begin{tabular}{lccccccccc}
\toprule
& \multicolumn{4}{c}{\textbf{\qwensmall}} & & \multicolumn{4}{c}{\textbf{\qwenmid}} \\
\cmidrule(lr){2-5}\cmidrule(lr){7-10}
& \multicolumn{2}{c}{CodeGen} & \multicolumn{2}{c}{SpecGen} & & \multicolumn{2}{c}{CodeGen} & \multicolumn{2}{c}{SpecGen} \\
\cmidrule(lr){2-3}\cmidrule(lr){4-5}\cmidrule(lr){7-8}\cmidrule(lr){9-10}
\textbf{Training Variants} & Pass@1 & Pass@10 & Pass@1 & Pass@10 & & Pass@1 & Pass@10 & Pass@1 & Pass@10 \\
\midrule
Base Model & 25.3 & 28.3 & 14.7 & 16.3 & & 39.7 & 53.5 & 29.4 & 41.7 \\
\midrule
SFT Spec-Only & 18.0 & 26.7 & 9.5 & 16.6 & & 41.5 & 68.4 & 30.4 & 50.0 \\
$\Delta$ Abs. (Rel.) & \textcolor{red}{$\downarrow$7.3 (-28.9\%)} & \textcolor{red}{$\downarrow$1.6 (-5.7\%)} & \textcolor{red}{$\downarrow$5.2 (-35.4\%)} & \textcolor{darkgreen}{$\uparrow$0.3 (+1.8\%)} & & \textcolor{darkgreen}{$\uparrow$1.8 (+4.5\%)} & \textcolor{darkgreen}{$\uparrow$14.9 (+27.9\%)} & \textcolor{darkgreen}{$\uparrow$1.0 (+3.4\%)} & \textcolor{darkgreen}{$\uparrow$8.3 (+19.9\%)} \\
\midrule
SFT Full & \textbf{43.5} & \textbf{72.7} & 22.8 & 38.8 & & \textbf{61.2} & 79.7 & 43.5 & 58.0 \\
$\Delta$ Abs. (Rel.) & \textcolor{darkgreen}{$\uparrow$18.2 (+71.9\%)} & \textcolor{darkgreen}{$\uparrow$44.4 (+156.9\%)} & \textcolor{darkgreen}{$\uparrow$8.1 (+55.1\%)} & \textcolor{darkgreen}{$\uparrow$22.5 (+138.0\%)} & & \textcolor{darkgreen}{$\uparrow$21.5 (+54.2\%)} & \textcolor{darkgreen}{$\uparrow$26.2 (+49.0\%)} & \textcolor{darkgreen}{$\uparrow$14.1 (+48.0\%)} & \textcolor{darkgreen}{$\uparrow$16.3 (+39.1\%)} \\
\midrule
SFT No-Test & 35.4 & 66.8 & \textbf{30.3} & \textbf{45.2} & & 58.7 & \textbf{84.0} & \textbf{47.6} & \textbf{63.9} \\
$\Delta$ Abs. (Rel.) & \textcolor{darkgreen}{$\uparrow$10.1 (+39.9\%)} & \textcolor{darkgreen}{$\uparrow$38.5 (+136.0\%)} & \textcolor{darkgreen}{$\uparrow$15.6 (+106.1\%)} & \textcolor{darkgreen}{$\uparrow$28.9 (+177.3\%)} & & \textcolor{darkgreen}{$\uparrow$19.0 (+47.9\%)} & \textcolor{darkgreen}{$\uparrow$30.5 (+57.0\%)} & \textcolor{darkgreen}{$\uparrow$18.2 (+61.9\%)} & \textcolor{darkgreen}{$\uparrow$22.2 (+53.2\%)} \\
\bottomrule
\end{tabular}
}
\vspace{-3mm}
\end{table*}


\smallsec{Setup.}
We evaluate on the \verina{} benchmark SpecGen task and compare with the official baseline.
The baseline uses 2-shot prompting and measures pass@1 averaged over 5 runs.
For \framework{}, we use the problem description and Lean signature provided by \verina{}.
The \verina{} test suite is private and not used in \framework{} synthesis.
We run the traceable refinement pipeline with up to 25 iterations and up to 60 tests generated per problems using LLM generated program as the test oracle, terminating early when a specification passes all internal tests and adversarial checks.
We evaluate 5 models: two closed-source models (Claude Opus 4.5, Claude Sonnet 4.5) and three open-weight models (\qwenlarge, \qwenmid, \qwensmall) to test whether improvements generalize across model families and scales.

\smallsec{Findings.}
Table~\ref{tab:verina_main} shows that \framework{} consistently improves specification synthesis across all evaluated models.
Claude Opus 4.5 achieves the strongest absolute performance at 86.6\% pass@1 (+27.6 points over baseline), while mid-tier open-weight models show the largest relative gains, with \qwenlarge improving by 86.7\% relative (+31.8 points).
These improvements generalize across model families, scales, and training paradigms (general purpose and code-specialized), demonstrating that traceable refinement provides robust benefits regardless of base model characteristics.

However, the magnitude of improvement correlates with model capability.
Mid-tier models (\qwenlarge, \qwenmid) gain 21.0--31.8 points, while the smallest model (\qwensmall) improves by only 7.2 points.
This pattern suggests that effectively exploiting requirement-level feedback requires sufficient reasoning capability to trace failures to specific constraints and perform targeted repairs.
Nevertheless, even weaker models benefit substantially in relative terms, and \Cref{sec:rq2} shows that traceable refinement substantially outperforms multiple baselines.

\subsection{RQ2: Which Components Enable Effective Refinement?}
\label{sec:rq2}

\smallsec{Setup.}
To isolate component contributions, we conduct ablations on two models representing different capability levels: Claude Opus 4.5 and \qwenmid.
All variants use the same iteration budget (25 refinements) and test generation capacity (60 tests per problem).

\smallsec{Ablations.}
We compare \framework{} against progressively structured baselines (\Cref{tab:verina_ablation}).
\textbf{Direct generation} is the \verina{} pass@1 baseline without any refinement.
\textbf{Direct refinement} adds iterative refinement using only LLM self-judgment: the model generates a specification, judges whether it satisfies the description, and revises for up to 25 iterations without test-based validation.
\textbf{Test-driven refinement} generates requirement-targeted tests and refines based on test outcomes, but without requirement decomposition or failure-attributed feedback.
\textbf{w/o adversarial testing} removes the adversarial test component only.
\textbf{\framework{}} includes all components described in~\Cref{sec:method}.

\smallsec{Findings.}
Table~\ref{tab:verina_ablation} reveals a clear hierarchy of component importance.
First, iteration without grounded validation is insufficient or harmful.
Direct refinement improves Opus 4.5 by 13.2 points but degrades \qwenmid by 2.3 points, demonstrating that naive LLM self-judgment can be counterproductive for weaker models.
Second, test-driven refinement without requirement attribution substantially underperforms.
While achieving 73.8\% on Opus 4.5, it only reaches 20.9\% on \qwenmid (worse than both direct refinement and baseline), demonstrating that test-based validation without proper attribution confuses weaker models.
The full pipeline outperforms this variant by 12.8 and 29.4 points respectively, indicating that grounded validation alone is insufficient without requirement-level attribution to guide localized repairs.
Third, removing adversarial testing reduces performance by 5.6 points on Opus 4.5 and 1.4 points on \qwenmid.
Adversarial testing effectively exposes under-constrained or incorrect specifications that pass requirement-targeted tests but miss edge cases, and this capability benefits more from stronger models' ability to generate diverse challenging cases.
In summary, effective refinement requires both grounded validation using test and requirement-level attribution.
The full \framework{} pipeline achieves consistent improvements across model scales.


\begin{table*}[ht]
\small
\centering
\caption{
\textbf{Transfer to out-of-domain benchmarks for \qwensmall}.
Linear merging ($\alpha{=}0.2$) successfully transfers specification-synthesis capabilities to math reasoning and general code generation tasks.
}
\label{tab:rq3_transfer}
\vspace{-2mm}
\resizebox{\textwidth}{!}{
\begin{tabular}{lccccccc}
\toprule
& \multicolumn{3}{c}{\textbf{Math Reasoning}} & \multicolumn{2}{c}{\textbf{Coding}} & \multicolumn{2}{c}{\textbf{\verina{}}} \\
\cmidrule(lr){2-4}\cmidrule(lr){5-6}\cmidrule(lr){7-8}
\textbf{Model variant} & \textbf{GSM8K} & \textbf{AIME24} & \textbf{AIME25} & \textbf{HumanEval} & \textbf{MBPP} & \textbf{CodeGen} & \textbf{SpecGen} \\
\midrule
Base Model & 66.3 & 60.0 & 46.7 & 63.7 & 75.2 & 25.3 & 14.7 \\
\midrule
Merged ($\alpha{=}0.2$) & 68.3 & \textbf{63.3} & 50.0 & \textbf{76.1} & \textbf{78.2} & 22.9 & 15.6 \\
$\Delta$ Abs. (Rel.) & \textcolor{darkgreen}{$\uparrow$2.0 (+3.0\%)} & \textcolor{darkgreen}{$\uparrow$3.3 (+5.5\%)} & \textcolor{darkgreen}{$\uparrow$3.3 (+7.1\%)} & \textcolor{darkgreen}{$\uparrow$12.4 (+19.5\%)} & \textcolor{darkgreen}{$\uparrow$3.0 (+4.0\%)} & \textcolor{red}{$\downarrow$2.4 (-9.5\%)} & \textcolor{darkgreen}{$\uparrow$0.9 (+6.1\%)} \\
\midrule
SFT No-Test & \textbf{80.7} & 40.0 & 23.3 & 70.1 & 70.2 & \textbf{35.4} & \textbf{30.3} \\
$\Delta$ Abs. (Rel.) & \textcolor{darkgreen}{$\uparrow$14.4 (+21.7\%)} & \textcolor{red}{$\downarrow$20.0 (-33.3\%)} & \textcolor{red}{$\downarrow$23.4 (-50.1\%)} & \textcolor{darkgreen}{$\uparrow$6.4 (+10.0\%)} & \textcolor{red}{$\downarrow$5.0 (-6.6\%)} & \textcolor{darkgreen}{$\uparrow$10.1 (+39.9\%)} & \textcolor{darkgreen}{$\uparrow$15.6 (+106.1\%)} \\
\bottomrule
\end{tabular}
}
\vspace{-3mm}
\end{table*}

\subsection{RQ3: Can Trajectory Distillation Improve Models?}
\label{sec:rq3}

\smallsec{Setup.}
We investigate whether refinement trajectories can be distilled into high-quality training data that improves base model capabilities.
We fine-tune \qwensmall and \qwenmid on variants of our dataset (\Cref{sec:dataset}) and evaluate the resulting models.
We assess improvement in two settings: in-domain performance on \verina{} CodeGen and SpecGen tasks, and out-of-domain transfer to general math and coding benchmarks.

\subsubsection{In-Domain Improvement: Lean Code and Specification Synthesis}

\smallsec{Training settings.}
We compare three variants derived from the same trajectory source.
\textit{SFT Spec-Only} uses only final specifications as targets, representing standard end-to-end NL-to-specification supervision as previous work suggested~\cite{cao2025informal}.
\textit{SFT Full} uses complete trajectory-distilled supervision spanning all tasks defined in~\Cref{sec:dataset}.
\textit{SFT No-Test} excludes test-rekated tasks from \textit{SFT Full}.
We train for 4 epochs on \qwensmall and 5 epochs on \qwenmid, using a learning rate of $2 \times 10^{-6}$ with a cosine schedule.
We evaluate fine-tuned models using \verina{} direct generation baselines on CodeGen and SpecGen tasks and report both pass@1 and pass@10, where pass@10 measures the model's ability to generate diverse candidates for downstream best-of-N selection~\cite{gui2024bondaligningllmsbestn} or reinforcement learning~\cite{setlur2025inferencetimefinetuning}.

\smallsec{Findings.}
Table~\ref{tab:rq3_training} shows that trajectory distillation substantially improves both tasks.
Notably, all models are trained on Lean specification synthesis trajectories only, yet we observe significant transfer gains to Lean code generation, demonstrating that reasoning about specifications benefits code synthesis as well.
\textit{SFT No-Test} achieves the best SpecGen performance, improving SpecGen pass@1 by 15.6--18.2 points across models, with particularly strong SpecGen@10 gains (+177\% on \qwensmall), indicating improved candidate distribution beyond greedy quality.
\textit{SFT Full} achieves the best CodeGen performance, improving CodeGen pass@1 by 18.2--21.5 points.

The supervision variant analysis reveals clear patterns.
First, \textit{SFT Spec-Only} fails despite using correct specifications as targets, providing minimal gains on \qwenmid and degrading \qwensmall.
This demonstrates that end-to-end supervision is likely insufficient without the intermediate reasoning steps.
Second, comparing \textit{SFT Full} and \textit{SFT No-Test} reveals that test-construction supervision primarily benefits code generation while specification synthesis benefits from focused training on decomposition and repair.
These results validate trajectory distillation as an effective approach, with process-level supervision substantially outperforming end-to-end examples and providing strong transfer to related tasks.

\subsubsection{Out-of-Domain: Transfer to General Benchmarks}

\smallsec{Training settings.}
To assess transfer to out-of-domain tasks, we evaluate the \qwensmall base model and the \textit{SFT No-Test} checkpoint from \Cref{sec:rq3} on general math and coding benchmarks: GSM8K~\cite{cobbe2021training}, AIME24~\cite{aime24}, AIME25~\cite{aime25}, HumanEval~\cite{chen2021evaluating}, and MBPP~\cite{austin2021program}.
To understand how trajectory data performs when mixed with general-purpose training, we apply linear model merging using merge-kit~\cite{goddard-etal-2024-arcees}, which interpolates parameters between the base and fine-tuned models.
We use an interpolation weight of $\alpha{=}0.2$ for the fine-tuned model, simulating training on a mixture of general and specification-synthesis data.

\smallsec{Findings.}
Table~\ref{tab:rq3_transfer} shows that trajectory-distilled training transfers to out-of-domain tasks.
The linear merged model (with $\alpha{=}0.2$) achieves consistent improvements across all out-of-domain benchmarks, with particularly strong gains on HumanEval, demonstrating that formal reasoning data benefits both math reasoning and code generation through improved structured problem-solving.
The full fine-tuned model on \textit{SFT No-Test} shows signs of overfitting to the specification-synthesis domain as expected.
While it strongly improves GSM8K, likely due to structured reasoning patterns transferring to step-by-step word problems, it degrades on competition mathematics.

We note that linear merging does not work reliably for \qwenmid due to its Mixture-of-Experts (MoE) architecture~\cite{zhou2025mergeme}, and we leave MoE-specific merging strategies to future work.


\section{Conclusion and Discussion}
\label{sec:conclusion}

We introduced \framework{}, a traceable refinement framework that synthesizes intent-aligned formal specifications through requirement-level attribution and localized repair, achieving 86.6\% pass@1 on \verina{} and establishing state-of-the-art results.
Ablation studies confirm that requirement decomposition and localized repair are critical for effective refinement.
Beyond inference-time gains, distilling \framework{}'s trajectories into 343K training data improves base model specification synthesis by 62--106\% while transferring to out-of-domain reasoning tasks, demonstrating that process-level supervision from traceable refinement teaches generalizable problem-solving strategies.


\smallsec{Future directions.}
While \framework{} advances state-of-the-art specification synthesis, several promising directions remain for future work.
First, exploring alternative oracle mechanisms beyond reference implementations like LLM-based prediction or human-in-the-loop feedback could enable specification synthesis in domains where trusted implementations are unavailable while maintaining intent-anchored validation.
Second, integrating advances in proof automation and emerging LLM-based provers promises to reduce reliance on LLM judge fallback for complex specifications, enabling more robust specification evaluation.
Third, we plan to extend our methodology beyond Lean 4 to other verification frameworks like Dafny~\cite{leino2010dafny} or Coq~\cite{barras1997coq} to demonstrate the generality of test-driven specification synthesis and broaden its impact across the formal methods community.

\section*{Impact Statement}

This paper presents work whose goal is to advance the field of machine learning. There are many potential societal consequences of our work, none of which we feel must be specifically highlighted here.

\bibliography{reference}
\bibliographystyle{icml2026}

\appendix

\onecolumn

\section{Dataset Statistics}
\label{app:dataset}

\subsection{Training Data Construction}

As detailed in~\Cref{sec:dataset}, we generate training data by executing \framework{} on TACO-verified problems~\citep{likaixin2024taco-verified} and distilling the resulting trajectories into supervised fine-tuning examples.
Starting from 12,901 decontaminated problems with zero 10-gram overlap with \verina{}, we successfully translate 11,981 to Lean and synthesize 6,842 specifications.

\subsection{Task Distribution}

The complete filtered dataset contains 343,827 examples spanning 11 distinct tasks extracted from \framework{}'s refinement pipeline.
Table~\ref{tab:task-distribution} shows the distribution of tasks in the full dataset.
Tasks fall into two categories: \emph{core specification tasks} that handle requirement decomposition, synthesis, and refinement; and \emph{test-related tasks} that generate and validate requirement-targeted tests.

\begin{table*}[t]
\centering
\small
\caption{Task distribution in the full SFT dataset.
Test-related tasks comprise 64.6\% of the dataset, while core specification tasks account for 35.4\%.}
\begin{tabular}{lrrl}
\toprule
\textbf{Task} & \textbf{Count} & \textbf{\%} & \textbf{Description} \\
\midrule
\multicolumn{4}{l}{\emph{Core Specification Tasks}} \\
\texttt{ar\_decomposition} & 6,927 & 2.0 & Decompose description into atomic requirements \\
\texttt{direct\_specgen} & 6,842 & 2.0 & End-to-end specification generation \\
\texttt{spec\_refinement} & 55,270 & 16.1 & Repair failed specifications with localized feedback \\
\texttt{feedback\_generation} & 42,330 & 12.3 & Generate requirement-attributed failure analysis \\
\texttt{signature\_generation} & 10,162 & 3.0 & Generate Lean function signatures \\
\midrule
\multicolumn{4}{l}{\emph{Test-Related Tasks}} \\
\texttt{positive\_test\_gen} & 8,054 & 2.3 & Generate positive tests (valid input-output pairs) \\
\texttt{negative\_input\_test\_gen} & 7,089 & 2.1 & Generate negative-input tests (invalid inputs) \\
\texttt{negative\_output\_test\_gen} & 7,612 & 2.2 & Generate negative-output tests (wrong outputs) \\
\texttt{adversarial\_test\_gen} & 1,759 & 0.5 & Generate adversarial tests for edge cases \\
\texttt{test\_mapping} & 79,639 & 23.2 & Map tests to atomic requirements \\
\texttt{verdict\_unknown} & 118,143 & 34.4 & Validate specifications on concrete test cases \\
\midrule
\textbf{Total} & \textbf{343,827} & \textbf{100.0} & \\
\bottomrule
\end{tabular}
\label{tab:task-distribution}
\end{table*}

\subsection{Training Variants}

To investigate the contribution of different task types, we construct three dataset variants from the same trajectory source.
Table~\ref{tab:training-variants} summarizes the three variants and their task composition.

\textbf{SFT Full} contains all 343,827 examples spanning the complete traceable refinement pipeline.
This variant provides comprehensive supervision for all components of VERISPECGEN, including requirement decomposition, test generation, specification synthesis, validation, and localized repair.

\textbf{SFT No-Test} excludes all six test-related tasks, retaining 121,531 examples focused on core specification synthesis.
We remove \texttt{positive\_test\_gen}, \texttt{negative\_input\_test\_gen}, \texttt{negative\_output\_test\_gen}, \texttt{adversarial\_test\_gen}, \texttt{test\_mapping}, and \texttt{verdict\_unknown}.
This variant concentrates supervision on requirement decomposition, specification generation, and requirement-attributed refinement.

\textbf{SFT Spec-Only} contains only 6,842 end-to-end examples from the \texttt{direct\_specgen} task, representing standard natural-language-to-specification supervision without intermediate reasoning steps.
This variant serves as a baseline to measure the value of process-level supervision.

\begin{table*}[t]
\centering
\small
\caption{Training dataset variants.
All variants are derived from the same high-quality checkpoints but differ in which tasks are included.}
\begin{tabular}{lrp{7cm}}
\toprule
\textbf{Variant} & \textbf{Examples} & \textbf{Description} \\
\midrule
SFT Full & 343,827 & All 11 tasks, complete refinement pipeline \\
SFT No-Test & 121,531 & 5 core specification tasks only \\
SFT Spec-Only & 6,842 & End-to-end NL-to-specification pairs only \\
\bottomrule
\end{tabular}
\label{tab:training-variants}
\end{table*}

\subsection{Task Composition by Variant}

Table~\ref{tab:task-by-variant} shows which tasks are included in each training variant and their relative proportions.
In SFT No-Test, refinement-related tasks (\texttt{spec\_refinement} and \texttt{feedback\_generation}) comprise 80.3\% of the dataset, providing concentrated supervision for requirement-attributed repair.
In contrast, SFT Spec-Only provides only final specification targets without any process-level supervision.

\begin{table*}[t]
\centering
\small
\caption{Task inclusion and percentage composition across training variants.
Percentages show the proportion of each task within its variant.
SFT No-Test concentrates supervision on refinement tasks (80.3\%), while SFT Spec-Only provides only end-to-end targets.}
\begin{tabular}{lcccc}
\toprule
\textbf{Task} & \textbf{Full} & \textbf{No-Test} & \textbf{Spec-Only} \\
\midrule
\texttt{direct\_specgen} & 2.0\% & 5.6\% & 100\% \\
\texttt{ar\_decomposition} & 2.0\% & 5.7\% & --- \\
\texttt{signature\_generation} & 3.0\% & 8.4\% & --- \\
\texttt{spec\_refinement} & 16.1\% & 45.5\% & --- \\
\texttt{feedback\_generation} & 12.3\% & 34.8\% & --- \\
\texttt{positive\_test\_gen} & 2.3\% & --- & --- \\
\texttt{negative\_input\_test\_gen} & 2.1\% & --- & --- \\
\texttt{negative\_output\_test\_gen} & 2.2\% & --- & --- \\
\texttt{adversarial\_test\_gen} & 0.5\% & --- & --- \\
\texttt{test\_mapping} & 23.2\% & --- & --- \\
\texttt{verdict\_unknown} & 34.4\% & --- & --- \\
\midrule
\textbf{Total Examples} & \textbf{343,827} & \textbf{121,531} & \textbf{6,842} \\
\bottomrule
\end{tabular}
\label{tab:task-by-variant}
\end{table*}

\subsection{Key Observations}

Our dataset construction reveals several insights about effective training for specification synthesis.

First, test-related tasks dominate the full dataset at 64.6\%, yet removing them (SFT No-Test) achieves the best specification synthesis performance (Table~\ref{tab:rq3_training}).
This suggests that focused training on core specification tasks is more effective than including test-generation supervision, possibly because test generation is a distinct capability that does not directly transfer to specification synthesis.

Second, end-to-end examples alone are insufficient for learning specification synthesis.
Despite containing correct final specifications as targets, SFT Spec-Only substantially underperforms both SFT Full and SFT No-Test.
This demonstrates that process-level supervision from intermediate reasoning steps is essential for teaching models how to achieve intent alignment.

Third, refinement supervision is critical for effective specification synthesis.
In SFT No-Test, \texttt{spec\_refinement} and \texttt{feedback\_generation} comprise 80.3\% of examples, providing explicit supervision for requirement-attributed repair.
The strong performance of models trained on SFT No-Test validates that learning to perform localized repairs based on requirement-level feedback is a key capability for specification synthesis.

\clearpage

\section{Additional Experiment Results}

\subsection{SFT Ablation Study}
\label{sec:sft_ablation}

We conduct a comprehensive ablation study to understand how training data composition affects specification generation performance.
We fine-tune both Qwen3-4B-Instruct-2507 and Qwen3-Coder-30B-A3B for 5 epochs and evaluate specification generation performance across sampling temperatures $T \in \{0.0, 0.3, 0.6, 1.0\}$ on the \verina{}| SpecGen task.

\subsubsection{Temperature Sensitivity}
\label{sec:temp_sensitivity}

\begin{figure}[t]
    \centering
    \includegraphics[width=\linewidth]{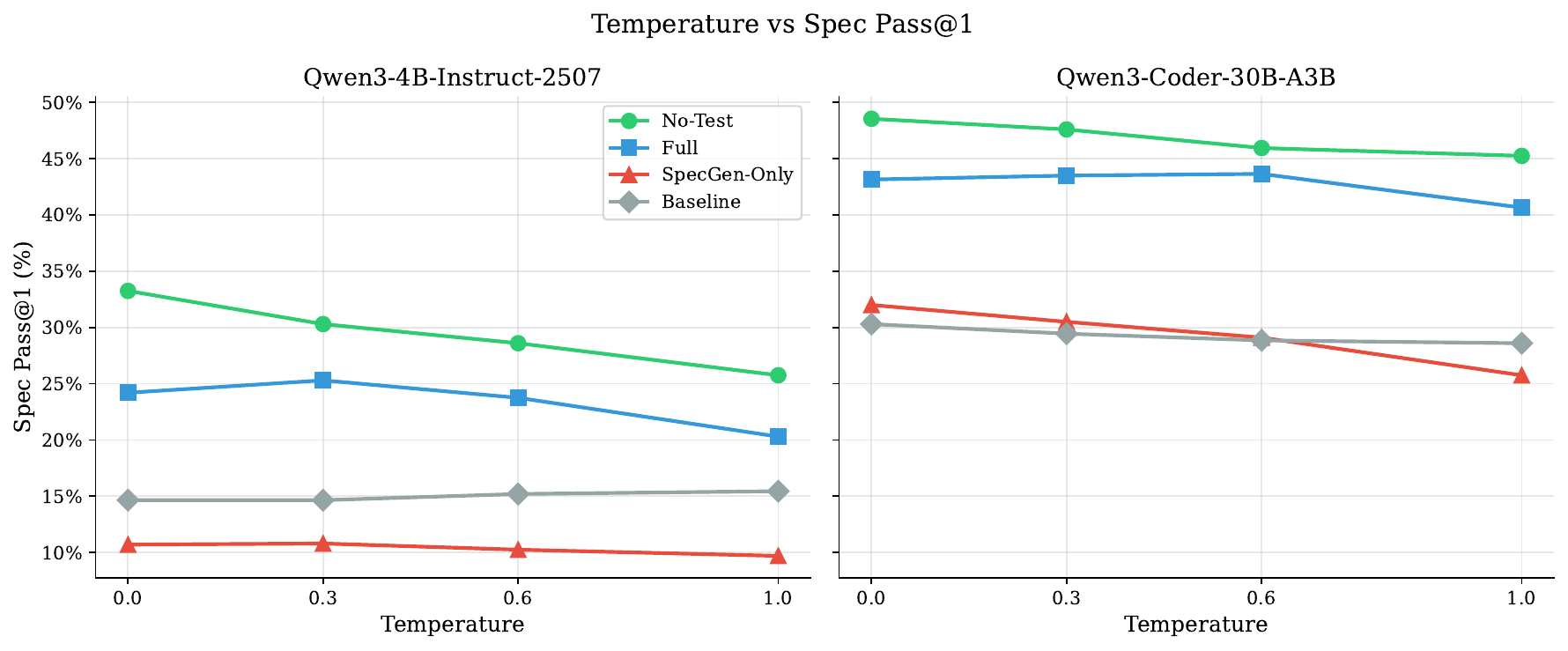}
    \caption{Temperature vs.\ Spec Pass@1 for different SFT configurations. Lower temperatures consistently yield better performance across all configurations. SFT Spec-Only drops below the base model at higher temperatures for the 30B model, demonstrating that single-task training provides insufficient learning signal and multi-task supervision from trajectory distillation is essential for robust specification synthesis.}
    \label{fig:temperature_sweep}
\end{figure}

\Cref{fig:temperature_sweep} shows the effect of sampling temperature on specification generation performance across training configurations.

\begin{itemize}
    \item \textbf{Lower temperature yields optimal performance}: Specification generation accuracy decreases monotonically with increasing temperature for all configurations. This pattern suggests that specification synthesis benefits from deterministic decoding, as greedy sampling ($T=0.0$) consistently achieves the best results. Models remain reasonably robust at moderate temperatures ($T \leq 0.6$), with performance degrading more sharply at $T=1.0$.
    
    \item \textbf{SFT No-Test consistently outperforms SFT Full}: Excluding test-related tasks improves specification generation by 2--5\% absolute across temperatures. This finding aligns with our observation in Section~5.3.1 that test-construction supervision primarily benefits code generation rather than specification synthesis, suggesting potential interference between test generation and specification generation learning objectives.
    
    \item \textbf{Multi-task training provides essential learning signal}: SFT Spec-Only, despite being trained specifically on specification generation, performs significantly worse than multi-task configurations. At high temperatures, the 30B model trained on SFT Spec-Only falls below the untrained base model, corroborating the finding in Table~\ref{tab:rq3_training} that end-to-end supervision without intermediate reasoning steps is insufficient for learning specification synthesis.
\end{itemize}

\subsubsection{Code vs.\ Specification Trade-off}
\label{sec:code_vs_spec}

\begin{figure}[t]
    \centering
    \includegraphics[width=\linewidth]{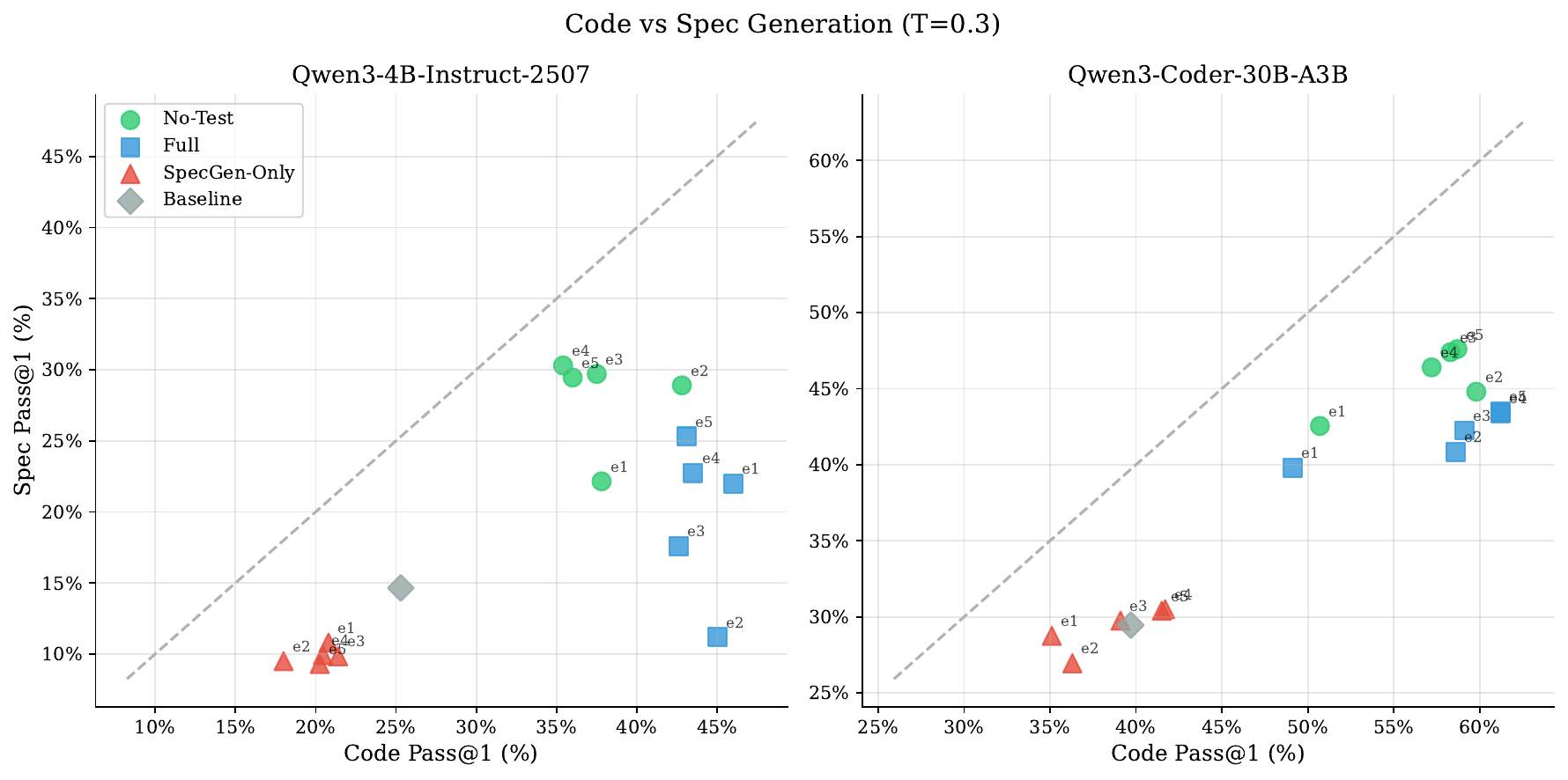}
    \caption{Code Pass@1 vs.\ Spec Pass@1 at $T=0.3$. Each point represents a training epoch (e1--e5). SFT No-Test achieves the best specification performance while maintaining competitive code generation capability. SFT Spec-Only clusters in the lower-left quadrant, exhibiting degraded performance on both tasks despite being trained specifically for specification generation.}
    \label{fig:code_vs_spec}
\end{figure}

\Cref{fig:code_vs_spec} visualizes the relationship between code generation and specification generation performance across training epochs. Points above the diagonal indicate configurations that achieve relatively stronger specification performance compared to code performance.

\begin{itemize}
    \item \textbf{SFT No-Test achieves optimal specification-code trade-off}: This configuration achieves the highest specification performance while maintaining code generation capability comparable to SFT Full, consistent with the results in Table~\ref{tab:rq3_training}.
    
    \item \textbf{Larger models exhibit clearer configuration separation}: The 30B model shows more distinct clustering between training configurations, suggesting that increased model capacity amplifies the effects of training data composition. This observation implies that careful curation of training tasks becomes increasingly important at larger scales.
\end{itemize}

\subsubsection{Training Dynamics Across Epochs and Temperatures}
\label{sec:epoch_temp_grid}

\begin{figure}[t]
    \centering
    \includegraphics[width=\linewidth]{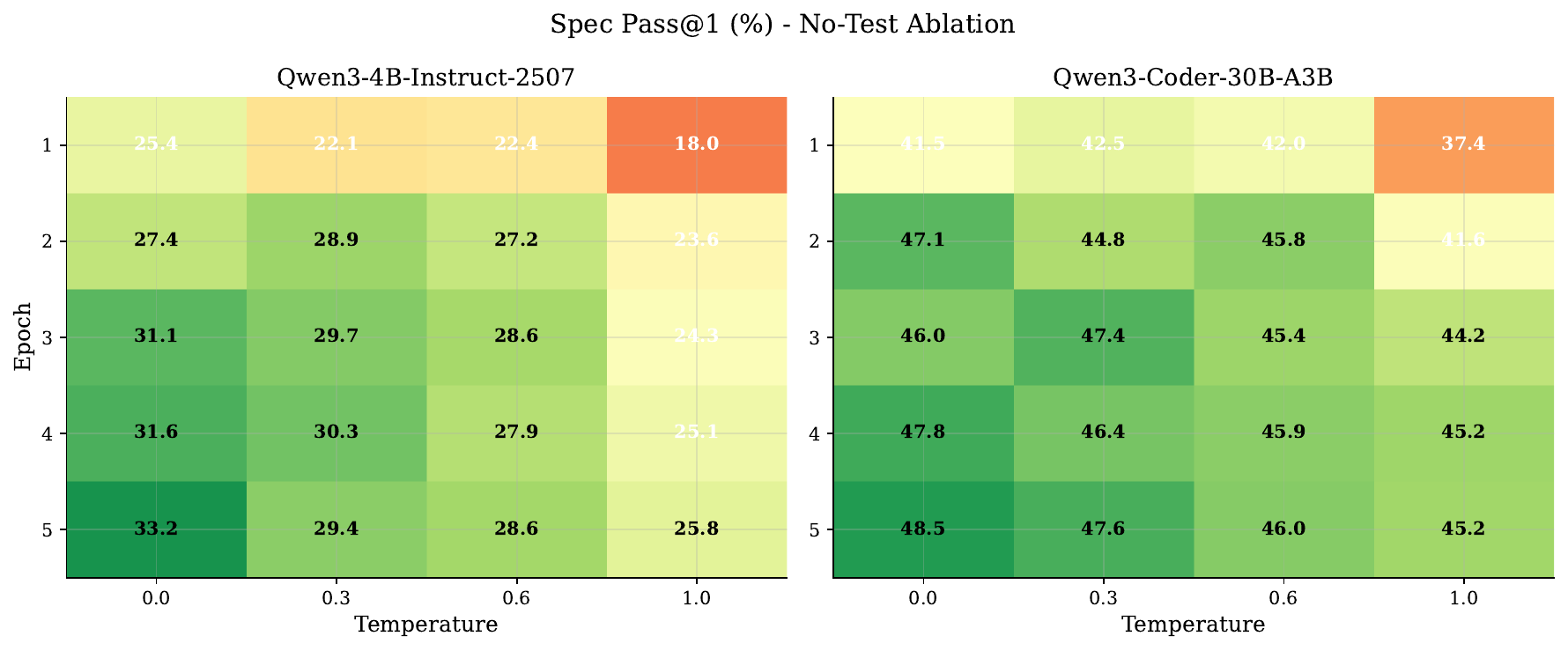}
    \caption{Spec Pass@1 (\%) across epochs and temperatures for SFT No-Test. Performance improves consistently with additional training epochs, with diminishing returns after epoch 3--4. At $T=0.3$, the best configuration achieves 30.3\% for the 4B model and 47.6\% for the 30B model, representing 15.6 and 18.2 points absolute improvement over the respective base models (14.7\% and 29.4\%).}
    \label{fig:heatmap}
\end{figure}

\Cref{fig:heatmap} presents a detailed analysis of how specification performance varies across training epochs and sampling temperatures for SFT No-Test, the best-performing configuration identified above.

\begin{itemize}
    \item \textbf{Consistent improvement with training epochs}: Both models show sustained improvement with additional training, though with diminishing returns after epoch 3--4. This suggests that 3--5 epochs provide sufficient exposure to the trajectory-distilled supervision without overfitting in-domain tasks.
\end{itemize}

\subsection{Full Out-of-Domain Transfer Results}
\label{app:transfer}

To assess whether trajectory-distilled training generalizes beyond formal verification, we evaluate \qwensmall{} on out-of-domain benchmarks as described in~\Cref{sec:rq3}.
We compare the base model, full fine-tuning on SFT No-Test, and linear model merging~\citep{xiao2024lm}, which interpolates parameters $\theta_{\text{merged}} = (1 - \alpha) \theta_{\text{base}} + \alpha \theta_{\text{ft}}$ to balance specialization with general capabilities.
We evaluate three interpolation weights $\alpha \in \{0.1, 0.2, 0.3\}$, where smaller $\alpha$ preserves base model capabilities and larger $\alpha$ incorporates more fine-tuned behavior.

Table~\ref{tab:transfer_full} shows that linear merging at $\alpha{=}0.2$ successfully transfers specification-synthesis capabilities to out-of-domain tasks, improving 6 of 7 benchmarks over the base model with particularly strong gains on HumanEval (+19.5\% relative).
The interpolation weight $\alpha$ controls the specialization-generalization trade-off: $\alpha{=}0.1$ achieves the best AIME25 (53.3) but minimal SpecGen gains (+0.8), while $\alpha{=}0.3$ improves SpecGen (+5.6) at the cost of MBPP degradation (-6.6\%).
Full fine-tuning (SFT No-Test) achieves the strongest in-domain performance but overfits to verification objectives.
Interestingly, GSM8K improves substantially (+21.7\%) even with full fine-tuning, likely because structured reasoning patterns in specification refinement transfer to step-by-step problem solving.
These results demonstrate that $\alpha{=}0.2$ merging provides a practical method to incorporate specification-synthesis capabilities into general-purpose models, while full fine-tuning is preferable for specialized verification applications.

\begin{table*}[ht]
\small
\centering
\caption{
\textbf{Transfer to out-of-domain benchmarks for \qwensmall}.
Linear merging with varying $\alpha$ values transfers specification-synthesis capabilities to math reasoning and general code generation tasks.
$\alpha{=}0.2$ achieves the best balance across out-of-domain benchmarks.
}
\label{tab:transfer_full}
\vspace{-2mm}
\resizebox{\textwidth}{!}{
\begin{tabular}{lccccccc}
\toprule
& \multicolumn{3}{c}{\textbf{Math Reasoning}} & \multicolumn{2}{c}{\textbf{Coding}} & \multicolumn{2}{c}{\textbf{\verina{}}} \\
\cmidrule(lr){2-4}\cmidrule(lr){5-6}\cmidrule(lr){7-8}
\textbf{Model variant} & \textbf{GSM8K} & \textbf{AIME24} & \textbf{AIME25} & \textbf{HumanEval} & \textbf{MBPP} & \textbf{CodeGen} & \textbf{SpecGen} \\
\midrule
Base Model & 66.3 & 60.0 & 46.7 & 63.7 & 75.2 & 25.3 & 14.7 \\
\midrule
Merged ($\alpha{=}0.1$) & 67.6 & 56.7 & \textbf{53.3} & 74.2 & 76.1 & 21.4 & 15.5 \\
$\Delta$ Abs. (Rel.) & \textcolor{darkgreen}{$\uparrow$1.3 (+2.0\%)} & \textcolor{red}{$\downarrow$3.3 (-5.5\%)} & \textcolor{darkgreen}{$\uparrow$6.6 (+14.1\%)} & \textcolor{darkgreen}{$\uparrow$10.5 (+16.5\%)} & \textcolor{darkgreen}{$\uparrow$0.9 (+1.2\%)} & \textcolor{red}{$\downarrow$3.9 (-15.4\%)} & \textcolor{darkgreen}{$\uparrow$0.8 (+5.4\%)} \\
\midrule
Merged ($\alpha{=}0.2$) & 68.3 & \textbf{63.3} & 50.0 & \textbf{76.1} & \textbf{78.2} & 22.9 & 15.6 \\
$\Delta$ Abs. (Rel.) & \textcolor{darkgreen}{$\uparrow$2.0 (+3.0\%)} & \textcolor{darkgreen}{$\uparrow$3.3 (+5.5\%)} & \textcolor{darkgreen}{$\uparrow$3.3 (+7.1\%)} & \textcolor{darkgreen}{$\uparrow$12.4 (+19.5\%)} & \textcolor{darkgreen}{$\uparrow$3.0 (+4.0\%)} & \textcolor{red}{$\downarrow$2.4 (-9.5\%)} & \textcolor{darkgreen}{$\uparrow$0.9 (+6.1\%)} \\
\midrule
Merged ($\alpha{=}0.3$) & 67.1 & 56.7 & 46.7 & 70.1 & 70.2 & 28.9 & 20.3 \\
$\Delta$ Abs. (Rel.) & \textcolor{darkgreen}{$\uparrow$0.8 (+1.2\%)} & \textcolor{red}{$\downarrow$3.3 (-5.5\%)} & \textcolor{black}{0.0 (+0.0\%)} & \textcolor{darkgreen}{$\uparrow$6.4 (+10.0\%)} & \textcolor{red}{$\downarrow$5.0 (-6.6\%)} & \textcolor{darkgreen}{$\uparrow$3.6 (+14.2\%)} & \textcolor{darkgreen}{$\uparrow$5.6 (+38.1\%)} \\
\midrule
SFT No-Test & \textbf{80.7} & 40.0 & 23.3 & 70.1 & 70.2 & \textbf{35.4} & \textbf{30.3} \\
$\Delta$ Abs. (Rel.) & \textcolor{darkgreen}{$\uparrow$14.4 (+21.7\%)} & \textcolor{red}{$\downarrow$20.0 (-33.3\%)} & \textcolor{red}{$\downarrow$23.4 (-50.1\%)} & \textcolor{darkgreen}{$\uparrow$6.4 (+10.0\%)} & \textcolor{red}{$\downarrow$5.0 (-6.6\%)} & \textcolor{darkgreen}{$\uparrow$10.1 (+39.9\%)} & \textcolor{darkgreen}{$\uparrow$15.6 (+106.1\%)} \\
\bottomrule
\end{tabular}
}
\vspace{-3mm}
\end{table*}

\clearpage

\DeclareTColorBox[auto counter,crefname={prompt}{prompts}]{prompt}{ o O{} t\label g }{%
    enhanced,
    IfValueTF={#1}{title={Prompt~\thetcbcounter\ (#1)}}{title=Prompt~\thetcbcounter}, IfBooleanTF={#3}{label=#4}{},#2}

\section{SFT Data Prompts}

We include the SFT dataset prompts and construction method from trajectories in the followings:

\begin{prompt}[AR Decomposition]
\textbf{System Prompt}

\begin{minipage}{\linewidth}
\begin{lstlisting}[language={}, breaklines=true, numbers=none, xleftmargin=0pt]
You are an expert in Lean 4 formal verification. Your task is to analyze
programming problems and identify the key requirements that a correct Lean 4
implementation must satisfy.

Requirements fall into two categories:
- **Preconditions**: Constraints on the input that must be true before the
function executes (e.g., "the list must not be empty", "n must be positive")
- **Postconditions**: Properties that the output must satisfy after the
function executes (e.g., "the result is the sum of all elements", "the
returned list is sorted")

Each requirement should be atomic, precise, and independently verifiable as
a Lean 4 proposition.
\end{lstlisting}
\end{minipage}

\tcbline
\textbf{User Prompt}

\begin{minipage}{\linewidth}
\begin{lstlisting}[language={}, breaklines=true, numbers=none, xleftmargin=0pt]
Analyze the following programming problem and identify all the requirements
(preconditions and postconditions) that a correct implementation must
satisfy.

## Problem Description
{problem_description}

## Function Signature
{signature}

Please identify all preconditions (input constraints) and postconditions
(output properties) for this problem. For each requirement, provide:
- A unique identifier (e.g., "R1", "R2")
- Whether it's a "precondition" or "postcondition"
- A clear description of the requirement
\end{lstlisting}
\end{minipage}

\tcbline
\textbf{Response (from trajectory data)}

Response uses \texttt{checkpoint.informal\_requirements} (the decomposed IRs), formatted as JSON with a natural language introduction stating the count of preconditions and postconditions.

\begin{minipage}{\linewidth}
\begin{lstlisting}[language={}, breaklines=true, numbers=none, xleftmargin=0pt]
I've analyzed the problem and identified {ir_count} requirements
({precond_count} preconditions and {postcond_count} postconditions).

```json
{requirements_json}
```
\end{lstlisting}
\end{minipage}

\end{prompt}

\begin{prompt}[Test Mapping]
\textbf{System Prompt}

\begin{minipage}{\linewidth}
\begin{lstlisting}[language={}, breaklines=true, numbers=none, xleftmargin=0pt]
You are an expert in Lean 4 formal verification. Your task is to analyze
test cases and determine which requirements they help verify for a Lean 4
specification.

For each requirement, you should assess:
- Whether the test case validates or covers this requirement
- Your confidence level in this assessment
- A brief explanation of your reasoning
\end{lstlisting}
\end{minipage}

\tcbline
\textbf{User Prompt}

\begin{minipage}{\linewidth}
\begin{lstlisting}[language={}, breaklines=true, numbers=none, xleftmargin=0pt]
Analyze which requirements this test case validates or covers.

## Requirements
{requirements_json}

## Test Case
Input: {test_input}
Expected Output: {test_output}

For each requirement, determine whether this test case validates it and
explain your reasoning.
\end{lstlisting}
\end{minipage}

\tcbline
\textbf{Response (from trajectory data)}

Response uses \texttt{checkpoint.testcase\_ar\_mapping\_detailed[tc.id]} which contains per-requirement mappings with \texttt{validates}, \texttt{confidence}, and \texttt{reason} fields for each test case.

\begin{minipage}{\linewidth}
\begin{lstlisting}[language={}, breaklines=true, numbers=none, xleftmargin=0pt]
I'll analyze which requirements this test case validates:

**{req_id}**: This test {status} this requirement ({confidence} confidence).
{reason}

```json
{detailed_mapping}
```
\end{lstlisting}
\end{minipage}

\end{prompt}

\begin{prompt}[Positive Test Generation]
\textbf{System Prompt}

\begin{minipage}{\linewidth}
\begin{lstlisting}[language={}, breaklines=true, numbers=none, xleftmargin=0pt]
You are an expert in Lean 4 software testing. Your task is to generate test
cases for programming problems that will be implemented in Lean 4.

**Positive test cases** are tests with valid inputs that satisfy all input
constraints (preconditions), paired with the correct expected output. These
tests verify that a correct Lean 4 implementation produces the right results
for valid inputs.
\end{lstlisting}
\end{minipage}

\tcbline
\textbf{User Prompt}

\begin{minipage}{\linewidth}
\begin{lstlisting}[language={}, breaklines=true, numbers=none, xleftmargin=0pt]
Generate positive test cases for the following problem. These should be
valid inputs with correct expected outputs.

## Problem Description
{problem_description}

## Function Signature
{signature}

## Requirements
{requirements_json}

Please generate test cases that cover different scenarios and edge cases.
For each test case, provide:
- The input values
- The expected output
- A brief description of what the test validates
\end{lstlisting}
\end{minipage}

\tcbline
\textbf{Response (from trajectory data)}

Response uses \texttt{checkpoint.testcases} formatted as JSON objects with \texttt{input}, \texttt{output}, and \texttt{description} fields.

\begin{minipage}{\linewidth}
\begin{lstlisting}[language={}, breaklines=true, numbers=none, xleftmargin=0pt]
I'll generate {test_count} positive test cases covering different scenarios
for this problem.

```json
{testcases_json}
```
\end{lstlisting}
\end{minipage}

\end{prompt}

\begin{prompt}[Negative Input Test Generation]
\textbf{System Prompt}

\begin{minipage}{\linewidth}
\begin{lstlisting}[language={}, breaklines=true, numbers=none, xleftmargin=0pt]
You are an expert in Lean 4 software testing. Your task is to generate test
cases for programming problems that will be implemented in Lean 4.

**Negative input test cases** are tests with invalid inputs that violate the
input constraints (preconditions). For example, if a function requires a
non-empty list, a negative input test would pass an empty list. These tests
help verify that a Lean 4 specification correctly identifies invalid inputs.
\end{lstlisting}
\end{minipage}

\tcbline
\textbf{User Prompt}

\begin{minipage}{\linewidth}
\begin{lstlisting}[language={}, breaklines=true, numbers=none, xleftmargin=0pt]
Generate negative input test cases for the following problem. These should
have invalid inputs that violate the input constraints.

## Problem Description
{problem_description}

## Function Signature
{signature}

## Requirements
{requirements_json}

Please generate test cases with invalid inputs that violate the
preconditions (input constraints). For each test case, provide:
- The invalid input values
- A description of which constraint is violated and why
\end{lstlisting}
\end{minipage}

\tcbline
\textbf{Response (from trajectory data)}

Response uses \texttt{checkpoint.negative\_input\_testcases} formatted as JSON objects with \texttt{input} and \texttt{description} fields.

\begin{minipage}{\linewidth}
\begin{lstlisting}[language={}, breaklines=true, numbers=none, xleftmargin=0pt]
I'll generate {test_count} negative input test cases that violate the input
constraints.

```json
{testcases_json}
```
\end{lstlisting}
\end{minipage}

\end{prompt}

\begin{prompt}[Negative Output Test Generation]
\textbf{System Prompt}

\begin{minipage}{\linewidth}
\begin{lstlisting}[language={}, breaklines=true, numbers=none, xleftmargin=0pt]
You are an expert in Lean 4 software testing. Your task is to generate test
cases for programming problems that will be implemented in Lean 4.

**Negative output test cases** are tests with valid inputs but intentionally
wrong outputs that violate the expected behavior (postconditions). For
example, if the correct output for input [1,2,3] is 6 (the sum), a negative
output test might use output 5 (incorrect sum). These tests help verify that
a Lean 4 specification correctly identifies wrong outputs.
\end{lstlisting}
\end{minipage}

\tcbline
\textbf{User Prompt}

\begin{minipage}{\linewidth}
\begin{lstlisting}[language={}, breaklines=true, numbers=none, xleftmargin=0pt]
Generate negative output test cases for the following problem. These should
have valid inputs but intentionally wrong outputs.

## Problem Description
{problem_description}

## Function Signature
{signature}

## Requirements
{requirements_json}

## Reference Test Cases (with correct outputs)
{positive_tests_json}

Please generate test cases by taking valid inputs and pairing them with
wrong outputs. For each test case, provide:
- The input values (valid inputs)
- A wrong output value that violates the expected behavior
- A description explaining why this output is wrong and which requirement it
violates
\end{lstlisting}
\end{minipage}

\tcbline
\textbf{Response (from trajectory data)}

Response uses \texttt{checkpoint.negative\_output\_testcases} formatted with \texttt{input}, \texttt{wrong\_output}, \texttt{correct\_output} (if available), and \texttt{description} fields.

\begin{minipage}{\linewidth}
\begin{lstlisting}[language={}, breaklines=true, numbers=none, xleftmargin=0pt]
I'll generate {test_count} negative output test cases with valid inputs but
wrong outputs.

```json
{testcases_json}
```
\end{lstlisting}
\end{minipage}

\end{prompt}

\begin{prompt}[Adversarial Test Generation]
\textbf{System Prompt}

\begin{minipage}{\linewidth}
\begin{lstlisting}[language={}, breaklines=true, numbers=none, xleftmargin=0pt]
You are an expert in Lean 4 software testing and verification. Your task is
to generate adversarial test cases designed to find potential bugs in Lean 4
specifications.

Adversarial tests are carefully crafted inputs that target:
- Edge cases and boundary conditions (e.g., empty arrays, maximum values,
zero)
- Corner cases that might be overlooked
- Inputs that might expose logical errors in the specification

These tests help ensure the Lean 4 specification is robust and correctly
handles all valid inputs.
\end{lstlisting}
\end{minipage}

\tcbline
\textbf{User Prompt}

\begin{minipage}{\linewidth}
\begin{lstlisting}[language={}, breaklines=true, numbers=none, xleftmargin=0pt]
Generate adversarial test cases to find potential bugs in this
specification.

## Problem Description
{problem_description}

## Function Signature
{signature}

## Requirements
{requirements_json}

## Current Specification
```lean
{spec}
```

Please generate test cases that target edge cases and boundary conditions.
Focus on inputs that might expose specification bugs. For each test, explain
why it might find a bug.
\end{lstlisting}
\end{minipage}

\tcbline
\textbf{Response (from trajectory data)}

Response uses \texttt{checkpoint.adversarial\_testcases} with \texttt{input}, \texttt{expected\_output}, and \texttt{description}. Metadata tracks if any test found a bug via \texttt{adversarial\_failure} flag.

\begin{minipage}{\linewidth}
\begin{lstlisting}[language={}, breaklines=true, numbers=none, xleftmargin=0pt]
I'll generate {test_count} adversarial test cases targeting edge cases and
boundary conditions that might expose specification bugs.

```json
{testcases_json}
```
\end{lstlisting}
\end{minipage}

\end{prompt}

\begin{prompt}[Specification Refinement]
\textbf{System Prompt}

\begin{minipage}{\linewidth}
\begin{lstlisting}[language={}, breaklines=true, numbers=none, xleftmargin=0pt]
You are a formal verification expert. Your task is to fix a Lean 4
specification based on the provided feedback.

The feedback describes issues found during evaluation:
- Syntax or type errors from Lean compilation
- Test case failures with specific conditions
- Logic errors in preconditions or postconditions

Generate a corrected specification that addresses all the feedback.
\end{lstlisting}
\end{minipage}

\tcbline
\textbf{User Prompt}

\begin{minipage}{\linewidth}
\begin{lstlisting}[language={}, breaklines=true, numbers=none, xleftmargin=0pt]
Fix this Lean 4 specification based on the feedback.

## Problem Description
{problem_description}

## Function Signature
{signature}

## Requirements
{requirements_json}

## Current (Failed) Specification
```lean
{failed_spec}
```

## Feedback
{feedback}

Generate a corrected Lean 4 specification with both precondition and
postcondition functions.
\end{lstlisting}
\end{minipage}

\tcbline
\textbf{Response (from trajectory data)}

Response uses \texttt{checkpoint.spec} (the final successful specification) as the corrected output. The input uses a failed attempt's spec and feedback from \texttt{checkpoint.spec\_attempts}.

\begin{minipage}{\linewidth}
\begin{lstlisting}[language={}, breaklines=true, numbers=none, xleftmargin=0pt]
Based on the feedback, I'll fix the specification to address the issues.

```lean
{corrected_spec}
```
\end{lstlisting}
\end{minipage}

\end{prompt}

\begin{prompt}[Feedback Generation]
\textbf{System Prompt}

\begin{minipage}{\linewidth}
\begin{lstlisting}[language={}, breaklines=true, numbers=none, xleftmargin=0pt]
You are a formal verification expert. Your task is to analyze a failed Lean
4 specification and generate detailed feedback explaining what's wrong and
how to fix it.

Your feedback should:
- Identify the root cause of failures
- Point to specific issues in preconditions or postconditions
- Suggest concrete fixes
- Be actionable and specific
\end{lstlisting}
\end{minipage}

\tcbline
\textbf{User Prompt}

\begin{minipage}{\linewidth}
\begin{lstlisting}[language={}, breaklines=true, numbers=none, xleftmargin=0pt]
Analyze this failed specification and generate feedback for improvement.

## Problem Description
{problem_description}

## Requirements
{requirements_json}

## Failed Specification
```lean
{failed_spec}
```

## Quality Score
{quality_score:.2f}

## Task
Generate detailed feedback explaining what's wrong and how to fix it.
\end{lstlisting}
\end{minipage}

\tcbline
\textbf{Response (from trajectory data)}

Response uses \texttt{attempt.informalized\_feedback} or \texttt{attempt.feedback} from \texttt{checkpoint.spec\_attempts}---the LLM-generated feedback explaining spec failures.

\begin{minipage}{\linewidth}
\begin{lstlisting}[language={}, breaklines=true, numbers=none, xleftmargin=0pt]
[Generated from checkpoint data as described above]
\end{lstlisting}
\end{minipage}

\end{prompt}

\begin{prompt}[Signature Generation]
\textbf{System Prompt}

\begin{minipage}{\linewidth}
\begin{lstlisting}[language={}, breaklines=true, numbers=none, xleftmargin=0pt]
You are a Lean 4 programming expert. Your task is to generate a function
signature in JSON format based on the problem description.

The signature should include:
- function_name: A descriptive name for the function
- input_params: A list of input parameters, each with a name and Lean 4 type
- output_type: The Lean 4 return type of the function
\end{lstlisting}
\end{minipage}

\tcbline
\textbf{User Prompt}

\begin{minipage}{\linewidth}
\begin{lstlisting}[language={}, breaklines=true, numbers=none, xleftmargin=0pt]
Generate a Lean 4 function signature for the following problem.

## Problem Description
{problem_description}

Output the signature as a JSON object with fields:
- function_name: string
- input_params: array of objects with "name" and "type" fields
- output_type: string
\end{lstlisting}
\end{minipage}

\tcbline
\textbf{Response (from trajectory data)}

Response uses \texttt{checkpoint.signature} (JSON format) with \texttt{function\_name}, \texttt{input\_params}, and \texttt{output\_type} fields.

\begin{minipage}{\linewidth}
\begin{lstlisting}[language={}, breaklines=true, numbers=none, xleftmargin=0pt]
Based on the problem description, I'll create a Lean 4 function signature.

The function `{func_name}` takes {num_params} parameter(s) and returns
`{output_type}`.

```json
{signature_json}
```
\end{lstlisting}
\end{minipage}

\end{prompt}

\begin{prompt}[Verdict Unknown]
\textbf{System Prompt}

\begin{minipage}{\linewidth}
\begin{lstlisting}[language={}, breaklines=true, numbers=none, xleftmargin=0pt]
You are a Lean 4 formal verification expert. Your task is to determine
whether a Lean 4 specification condition will evaluate to TRUE or FALSE for
a given test case.

Analyze the Lean 4 specification logic carefully and determine the
evaluation result.
Provide clear reasoning for your conclusion.
\end{lstlisting}
\end{minipage}

\tcbline
\textbf{User Prompt}

\begin{minipage}{\linewidth}
\begin{lstlisting}[language={}, breaklines=true, numbers=none, xleftmargin=0pt]
Determine if this {component} will evaluate to TRUE or FALSE.

## Specification
```lean
{spec}
```

## Test Case
Type: {test_type}
Input: {test_input}
Output: {test_output}

**Test Type Explanation:**
- **positive**: Valid inputs that should produce correct outputs
- **neg_input**: Invalid inputs that violate preconditions (should fail
precondition check)
- **neg_output**: Valid inputs with wrong outputs (should pass precondition
but fail postcondition)

## Task
Analyze whether the {component} condition evaluates to TRUE or FALSE for
this test case. Explain your reasoning step by step, then provide your final
answer.
\end{lstlisting}
\end{minipage}

\tcbline
\textbf{Response (from trajectory data)}

Response uses \texttt{result.llm\_reasoning} from \texttt{checkpoint.spec\_eval.results} where \texttt{result\_status=`unknown'}, combined with the computed TRUE/FALSE verdict.

\begin{minipage}{\linewidth}
\begin{lstlisting}[language={}, breaklines=true, numbers=none, xleftmargin=0pt]
Let me analyze the {component} for this test case.

{reasoning}

**Answer: {eval_result}**
\end{lstlisting}
\end{minipage}

\end{prompt}

\begin{prompt}[Direct Specification Generation]
\textbf{System Prompt}

\begin{minipage}{\linewidth}
\begin{lstlisting}[language={}, breaklines=true, numbers=none, xleftmargin=0pt]
Your input fields are:
1. `task_description` (str): The specification task
2. `task_template` (str): Lean 4 code snippet with placeholders
3. `precond_desc` (str): Natural language precondition description
4. `postcond_desc` (str): Natural language postcondition description
Your output fields are:
1. `imports` (str): Required imports (optional)
2. `precond_aux` (str): Auxiliary precondition definitions
3. `precond` (str): Generated precondition code
4. `postcond_aux` (str): Auxiliary postcondition definitions
5. `postcond` (str): Generated postcondition code
All interactions will be structured in the following way, with the
appropriate values filled in.

[[ ## task_description ## ]]
{task_description}
[[ ## task_template ## ]]
{task_template}
[[ ## precond_desc ## ]]
{precond_desc}
[[ ## postcond_desc ## ]]
{postcond_desc}
[[ ## imports ## ]]
{imports}
[[ ## precond_aux ## ]]
{precond_aux}
[[ ## precond ## ]]
{precond}
[[ ## postcond_aux ## ]]
{postcond_aux}
[[ ## postcond ## ]]
{postcond}
[[ ## completed ## ]]

In adhering to this structure, your objective is:
        You are an expert in Lean 4 programming and theorem proving.
        Please generate a Lean 4 specification that constrains the program
implementation using the template provided in `task_template`.
        The `task_template` is a Lean 4 code snippet that contains
placeholders (wrapped with {{}}) for the spec to be generated.
        The precondition should be as permissive as possible, and the
postcondition should model a sound and complete relationship between input
and output of the program based on the `task_description`.
        The generated specification should:
        - Be well-documented with comments if necessary
        - Follow Lean 4 best practices and use appropriate Lean 4 syntax and
features
        - DO NOT use Lean 3 syntax or features
        - DO NOT import Std or Init
        - Only use `precond_aux` or `postcond_aux` when you cannot express
the precondition or postcondition in the main body of the specification
        - add @[reducible, simp] attribute to the definitions in
`precond_aux` or `postcond_aux`
        Hint:
        - Use a[i]! instead of a[i] when a is an array or a list when
necessary
\end{lstlisting}
\end{minipage}

\end{prompt}

\begin{prompt}[Direct Specification Generation (Continued)]
\textbf{User Prompt}

\begin{minipage}{\linewidth}
\begin{lstlisting}[language={}, breaklines=true, numbers=none, xleftmargin=0pt]
[[ ## task_description ## ]]
{task_description}

[[ ## task_template ## ]]
{task_template}

[[ ## precond_desc ## ]]
{precond_desc}

[[ ## postcond_desc ## ]]
{postcond_desc}

Respond with the corresponding output fields, starting with the field `[[ ##
imports ## ]]`, then `[[ ## precond_aux ## ]]`, then `[[ ## precond ## ]]`,
then `[[ ## postcond_aux ## ]]`, then `[[ ## postcond ## ]]`, and then
ending with the marker for `[[ ## completed ## ]]`.
\end{lstlisting}
\end{minipage}

\tcbline
\textbf{Response (from trajectory data)}

Response uses parsed components from \texttt{checkpoint.spec}: \texttt{imports}, \texttt{precond\_aux}, \texttt{precond}, \texttt{postcond\_aux}, \texttt{postcond} extracted via \texttt{parse\_spec\_components()}.

\begin{minipage}{\linewidth}
\begin{lstlisting}[language={}, breaklines=true, numbers=none, xleftmargin=0pt]
[[ ## imports ## ]]
{imports}

[[ ## precond_aux ## ]]
{precond_aux}

[[ ## precond ## ]]
{precond}

[[ ## postcond_aux ## ]]
{postcond_aux}

[[ ## postcond ## ]]
{postcond}

[[ ## completed ## ]]
\end{lstlisting}
\end{minipage}

\end{prompt}

\clearpage

\end{document}